\documentclass[manuscript,nonacm,10pt,twocolumn]{acmart}

\renewcommand\footnotetextcopyrightpermission[1]{}
\usepackage[table]{xcolor} 
\definecolor{avgblue}{RGB}{230,242,255}
\usepackage{microtype}
\usepackage{graphicx}
\usepackage{subcaption}
\usepackage{booktabs}
\usepackage{multirow}
\usepackage{tabularx}
\usepackage{makecell}
\usepackage{mathtools}
\usepackage{amsfonts}

\usepackage{algorithm}
\usepackage{algpseudocode}

\usepackage[capitalize,noabbrev]{cleveref}

\usepackage{float}     
\usepackage{placeins}  

\usepackage{xspace}
\DeclareRobustCommand{\mymodel}{\mbox{H$^{2}$MT}\xspace}

\makeatletter
\let\ACM@origbaselinestretch\baselinestretch
\patchcmd{\@printtopmatter}{\box\mktitle@bx\par}{\twocolumn[\box\mktitle@bx]}%
  {\typeout{H2MT: title-block twocolumn patch OK}}%
  {\typeout{H2MT: title-block twocolumn patch FAILED}}
\authorsaddresses{}
\def\@mkabstract{\bgroup
  \ifx\@abstract\@lempty\else
  {\phantomsection\addcontentsline{toc}{section}{\abstractname}%
    \section*{\abstractname}%
    \small
   \ignorespaces\@abstract\par}%
  \fi\egroup}
\makeatother
\AtBeginDocument{}

\begin{document}

\title[H$^{2}$MT]{H$^{2}$MT: Semantic Hierarchy-Aware Hierarchical Memory Transformer}

\author{Maryam Haghifam}
\affiliation{%
  \institution{University of California, Los Angeles}
  \city{Los Angeles}
  \country{USA}}
\email{maryamhgf@cs.ucla.edu}

\author{Zifan He}
\affiliation{%
  \institution{University of California, Los Angeles}
  \city{Los Angeles}
  \country{USA}}
\email{zifanhe@cs.ucla.edu}

\author{Jason Cong}
\affiliation{%
  \institution{University of California, Los Angeles}
  \city{Los Angeles}
  \country{USA}}
\email{cong@cs.ucla.edu}

\author{Yizhou Sun}
\affiliation{%
  \institution{University of California, Los Angeles}
  \city{Los Angeles}
  \country{USA}}
\email{yzsun@cs.ucla.edu}

\begin{abstract}
Transformer-based LLMs achieve strong results on many language tasks; however, long inputs remain challenging because context windows are finite and prefill latency and memory grow rapidly with prompt length. Flat token-stream processing and chunk-based retrieval can therefore spend substantial computation and context budget on text unrelated to the query. Offline-indexed RAG additionally introduces external storage and index management overhead, and typically appends retrieved evidence as raw text, increasing prefill cost and latency.
H$^{2}$MT makes long-context inference structure-aware: it builds a semantic hierarchy offline, computes a memory embedding for each node via bottom-up post-order aggregation, and routes queries coarse-to-fine at inference to prune irrelevant branches early. On LongBench QA (NarrativeQA, HotpotQA, QASPER) and the public
OpenROAD technical-document benchmark, H$^{2}$MT achieves favorable
quality--efficiency trade-offs, with lower average TTFT and peak GPU memory than HMT across the evaluated backbones, and substantially lower offline construction cost than hierarchical retrieval. The gains are strongest when documents provide meaningful native
structure; in our weak-structure HotpotQA comparison, raw-text
retrieval remains stronger.
\end{abstract}

\keywords{long-context inference, memory tokens, hierarchical routing, efficient LLM inference}

\maketitle


\begin{figure*}[t]
  \centering
  \includegraphics[width=\textwidth]{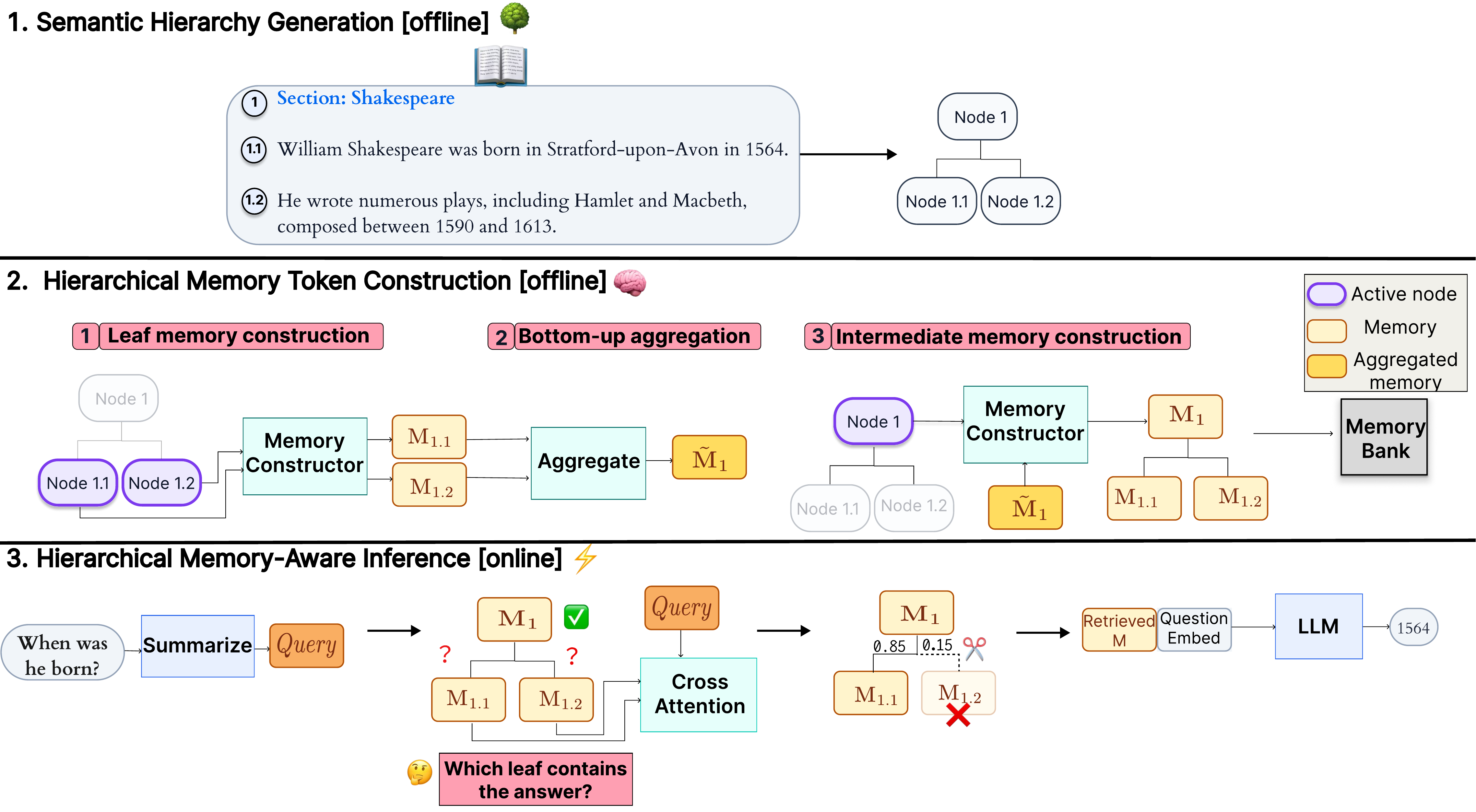}
    \caption{Overview of H$^{2}$MT.
    (1) Semantic hierarchy construction (offline): represent a structured
    document as a rooted tree of semantically coherent units.
    (2) Hierarchical memory construction (offline): compute leaf memories
    using learnable memory-marker embeddings and propagate information
    bottom-up by aggregating child memories into internal-node memories.
    (3) Hierarchical memory-aware inference (online): encode the question,
    route top-down through the hierarchy, and condition generation on the
    selected node memories.}
  \label{fig:main}
\end{figure*}

\section{Introduction}
Many applications repeatedly query the same long document or document collection, including scientific papers, technical manuals, legal codes, and software documentation. Such inputs are costly for models: without reusable document state, every query repeats the prefill computation, whose cost grows quadratically with sequence length under standard dense attention \citep{vaswani2017attention}, and very long inputs may exceed the model's context window entirely. Moreover, a query may depend on only a small region of the document, and language models do not reliably use relevant evidence when it appears within a long input \citep{liu2024lost}. The central problem is therefore not only how to support and accelerate long-context inference, but how to organize reusable document state compactly so that relevant information can be selected efficiently for each query.

Existing approaches address different components of this problem. Prompt-compression and latent-memory methods shorten the context length \citep{jiang-etal-2024-longllmlingua,ge2024incontext,he-etal-2025-hmt}, while sparse and approximate attention methods reduce the cost of processing long token sequences directly \citep{beltagy2020longformer,zaheer2020bigbird,wang2020linformer}. Retrieval-augmented generation supplies selected text to the reader \citep{sarthi2024raptor}, whereas selective-attention and KV-cache systems reduce the storage or access cost of token-level states \citep{retroinfer,arkvale}. Most of these methods still organize context as tokens, segments, or retrieved chunks. Among the closest structure-aware alternatives, HDT modifies the backbone attention pattern to incorporate hierarchy \citep{hdt}, while RAPTOR retrieves text and generated summaries from an induced hierarchy \citep{sarthi2024raptor}. Neither combines document hierarchy with compact latent node memories and query-dependent coarse-to-fine routing.

This creates a mismatch with real-world artifacts, whose semantic
organization, including sections, subsections, and scoped definitions,
provides a useful inductive bias for locating and composing evidence.
Ignoring this organization can allocate computation and context budget
to distant or irrelevant content, particularly in technical documents
with localized terminology and hierarchical dependencies.

We propose H$^{2}$MT, a hierarchy-aware latent-memory framework
that preserves a document's semantic hierarchy and moves
document-level conditioning from raw-token prefill to
query-conditioned selection over composed latent memories.
H$^{2}$MT represents available document organization, such as section
or table-of-contents structure, as a rooted semantic tree. It computes
one fixed-size memory embedding per node, distilling the node's local
content and descendant information through bottom-up aggregation.
At inference time, H$^{2}$MT routes the query coarse-to-fine from the
root, prunes low-scoring subtrees early, and conditions generation on
a bounded set of selected node memories rather than on the raw
document. Because selected nodes may occur at different semantic levels, the
per-parent routing budget controls both the amount and granularity of
the memory context supplied to the reader model. Consequently, the reader-side context length depends on the number
of selected nodes rather than on the raw-text length of those document
regions. The pretrained backbone remains frozen; LoRA adapters,
learnable memory-marker embeddings, lightweight aggregation modules,
and routing projections implement the trainable memory interface
without changing the backbone attention architecture.
Because node memories are document-dependent but query-independent,
they are constructed once and reused across questions. At query time,
H$^{2}$MT therefore performs only question encoding, hierarchical
routing, and generation over the selected memories, rather than
reprocessing the complete document.
H$^{2}$MT is designed primarily for repeated question answering over documents with meaningful semantic hierarchy. We additionally study GMM-based hierarchy induction as an exploratory extension for weakly structured contexts. We therefore treat weakly structured multi-document question answering as a stress test rather than as the primary evidence for the method.

Our contributions are as follows. (1) We introduce H$^{2}$MT, 	a hierarchy-organized latent-memory framework that caches one fixed-size memory embedding per node of a document hierarchy while leaving the backbone attention architecture unchanged. (2) We develop offline bottom-up node-memory construction and online top-down routing with per-parent pruning, producing a bounded query-specific memory context without reprocessing the raw document. (3) We evaluate answer quality, TTFT, peak GPU memory, routing-budget
sensitivity, and repeated-query amortization across LongBench QA and
the public OpenROAD technical-document benchmark. We compare against latent-memory, prompt-compression, and hierarchical retrieval baselines and provide auxiliary comparisons with selective-attention and KV-cache systems, identifying both the structured settings in which H$^{2}$MT provides favorable trade-offs and the weak-structure settings in which raw-text retrieval remains stronger.


\section{Related Work}
\paragraph{Structure-aware and memory-based models.} The Hierarchical Document Transformer (HDT) \citep{hdt} incorporates document structure through anchor tokens, sparse attention, and hierarchical positional encodings. H$^{2}$MT instead keeps the backbone attention architecture unchanged and uses the hierarchy as an addressing structure over cached latent node memories. HMT \citep{he-etal-2025-hmt} compresses sequential segments into recurrent multi-tier memories, while Memorizing Transformers \citep{wu2022memorizing} retrieve cached activations through kNN lookup. In contrast, H$^{2}$MT organizes memories according to semantic containment rather than sequence order or a flat activation store.

\paragraph{Efficient attention and context compression.} Longformer \citep{beltagy2020longformer}, Performer \citep{choromanski2021rethinking}, Transformer-XL \citep{dai2019transformerxl}, and RMT \citep{bulatov2022rmt} reduce long-sequence cost through sparse, approximate, or recurrent computation. ICAE \citep{ge2024incontext}, R$^{3}$Mem \citep{wang-etal-2025-r3mem}, and LongLLMLingua \citep{jiang-etal-2024-longllmlingua} compress long inputs into latent representations or shorter prompts. These approaches generally operate over flat token or segment sequences, whereas H$^{2}$MT organizes latent memories according to document hierarchy.

\paragraph{Hierarchical retrieval and KV-cache systems.} RAPTOR \citep{sarthi2024raptor} recursively clusters text and generates summaries for multi-level retrieval. H$^{2}$MT uses a related recursive construction only when native structure is unavailable, but routes over fixed-size latent memories rather than supplying retrieved text to the reader. RetroInfer \citep{retroinfer}, ArkVale \citep{arkvale}, MagicPIG \citep{magicpig}, and IceCache \citep{icecache} instead optimize access to token-level KV states. These methods act at a different inference stage and are
complementary to hierarchy-organized memory selection.


\section{Methodology}
H$^{2}$MT separates document-dependent computation from query-dependent inference. Given a static document or corpus, the offline stage constructs a semantic hierarchy and computes one reusable latent memory for each node. For each subsequent query, the online stage encodes the question, routes over the cached hierarchy, and conditions generation on a bounded set of selected node memories. We first formalize this repeated-query setting and its cost model, then describe hierarchy construction, hierarchical memory construction, query-time routing, and training.

\subsection{Problem Setting and Cost Model}

Let $D$ denote a static long document or document corpus that is queried repeatedly through a sequence of questions $\mathcal{Q}=\{q_i\}_{i=1}^{N}$. We represent $D$ as a rooted semantic tree $\mathcal{T}=(V,E,r)$, where each node $v\in V$ corresponds to a semantically coherent document unit and is associated with local text $x_v$. H$^{2}$MT constructs one fixed-size latent memory $m_v\in\mathbb{R}^{d}$ for every node and caches the memory bank $\mathcal{M}=\{m_v\}_{v\in V}$ for reuse across queries.

We separate document-dependent offline cost from query-dependent online cost. Let \(T_{\mathrm{off}}=T_{\mathrm{hier}}+T_{\mathrm{mem}}\) denote one-time hierarchy and node-memory construction, and let \(T_{\mathrm{online}}(q_i)\) denote query encoding, routing, reader prefill, and generation for query \(q_i\). After \(N\) queries, the average cost is \(\overline{T}(N)=T_{\mathrm{off}}/N+\frac{1}{N}\sum_{i=1}^{N}T_{\mathrm{online}}(q_i)\), making the one-time document cost explicit.

\subsection{Semantic Hierarchy Construction}
\label{sec:semantic_tree}
The edges of $\mathcal{T}$ encode relations between semantically coherent document units, such as sections, subsections, and paragraphs. Let
\(p: V\setminus\{r\}\to V\) denote the parent map and
\(C(v)\triangleq \{u\in V: p(u)=v\}\) the set of children of \(v\).
We write \(\mathcal{T}_v\) for the subtree rooted at \(v\).
Each node \(v\) is associated with local text \(x_v\), which fits within the backbone model context window. If a unit exceeds the backbone context window, we recursively split it into paragraphs and then into fixed-length chunks. When an explicit document semantic hierarchy is available (as in technical documents), we preserve its structure rather than inducing a new hierarchy. When reliable structure is unavailable, we optionally construct a hierarchy through recursive clustering over node memories. Dataset-specific hierarchy definitions and search scopes are reported in Section~\ref{sec:experimental-setup}.

\begin{figure}[t]
  \centering
  \includegraphics[width=\columnwidth]{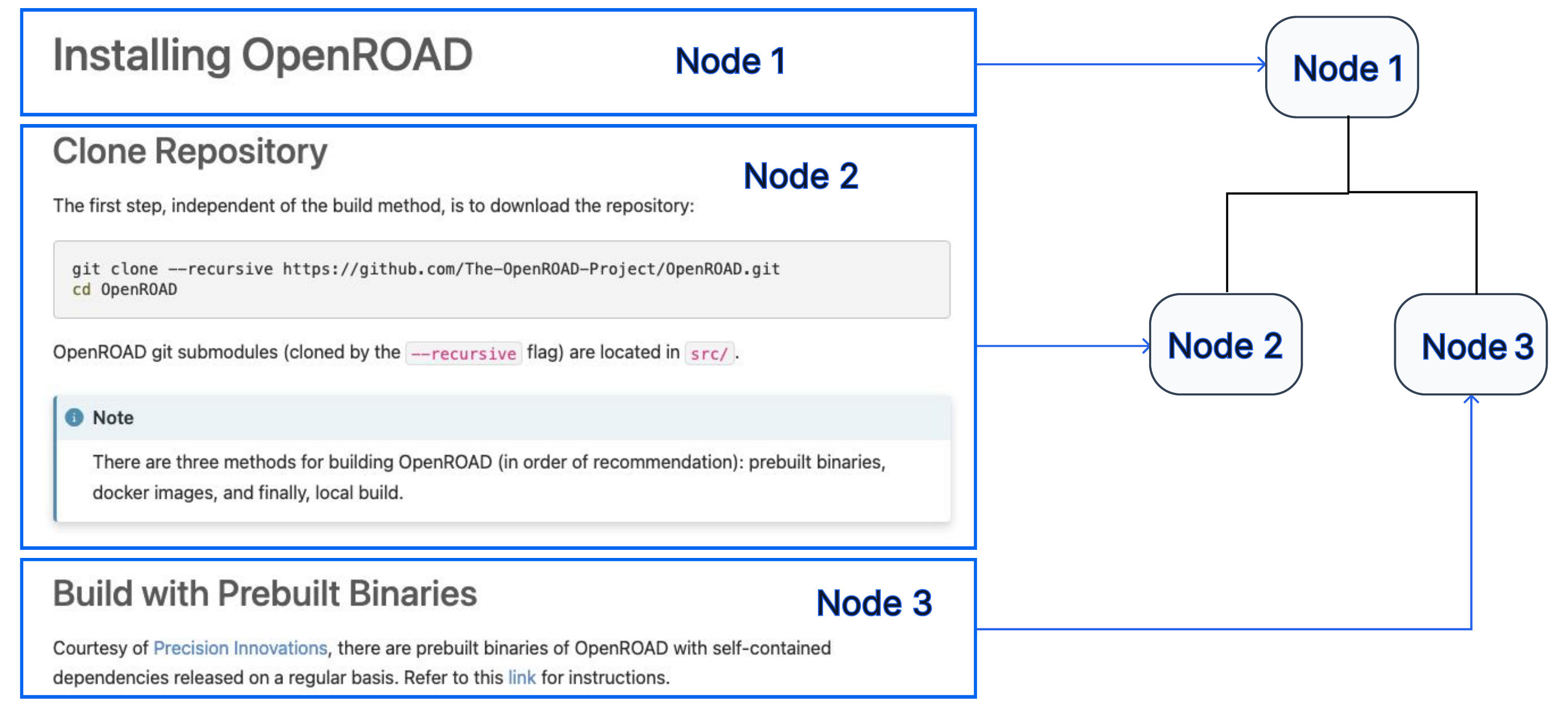}
  \caption{Technical document example.}
  \label{fig:or_example}
\end{figure}

\subsection{Hierarchical Memory Construction}
\label{sec:bottom_up}
We describe memory construction for leaf and internal nodes and formalize how information is propagated bottom-up through the hierarchy. We use a post-order traversal so that all descendants of an internal node are
processed before the node itself. Document-side memory construction is performed offline; all node memories
$\{m_v\}_{v\in V}$ can be cached and reused across queries.
H$^{2}$MT does not modify the backbone attention architecture or its causal attention mask, and the pretrained base-model weights remain
frozen.

The trainable components differ between stages. Corpus training updates
the LoRA adapters, write/read embedding parameters, and the selected
aggregation module. During QA fine-tuning, cached document-side memories
and aggregation parameters are fixed, while the generation and routing
components are optimized through their corresponding losses. The trained
LoRA adapter is specialized to the latent-memory interface rather than
ordinary text conditioning; Appendix~\ref{app:lora-plain-text} reports
the corresponding plain-text ablation.

\subsubsection{Leaf Memory Construction}
\label{sec:mem_emb}
We construct a single memory embedding for each node \(v\), denoted \(m_v \in \mathbb{R}^d\).
For a leaf node, \(m_v\) is produced by augmenting the backbone input with learnable write and read memory-marker embeddings, denoted \(e_{\mathrm{write}},e_{\mathrm{read}}\in\mathbb{R}^{d}\). These marker embeddings are shared across all nodes. Their parameter-sharing configuration and initialization are reported in Appendix~\ref{app:hyper}. We form the augmented embeddings
\[
\tilde{E}_v \;=\; [\, e_{\mathrm{write}} \,;\; E_v \,;\; e_{\mathrm{read}} \,]
\in \mathbb{R}^{(n_v+2)\times d}.
\]
We run the \(L\)-layer causal Transformer backbone on \(\tilde{E}_v\). Under the causal attention mask, the final position corresponding to \(e_{\mathrm{read}}\) can attend to all preceding positions and therefore acts as a learned readout slot that compresses the leaf content.

\subsubsection{Internal-Node Memory Construction}
\label{sec:mem_emb2}
For an internal node \(v\), \(m_v\) should reflect both its local text \(x_v\) and the information contained in its descendants. Naively concatenating raw descendant text to \(x_v\) can exceed the backbone context window. Instead, we propagate compressed information upward using the descendant node memories.

Let \(v\in V\) be an internal node, let \(c_v \triangleq |C(v)|\), and enumerate its children in document order as \(C(v)=(u_1,\ldots,u_{c_v})\). We stack their memories row-wise:
\[
M^{\mathrm{ch}}_v
\;\triangleq\;
\begin{bmatrix}
m_{u_1}^{\top}\\
\vdots\\
m_{u_{c_v}}^{\top}
\end{bmatrix}
\in \mathbb{R}^{c_v\times d}.
\]

We instantiate $\mathrm{Agg}$ using lightweight aggregation policies that summarize the child memories into one vector: mean pooling, self-attention, parent-conditioned cross-attention, GAT-style weighting, or parent-token self-attention. The cross-attention and GAT variants use parent-side representations as conditioning inputs, whereas parent-token self-attention prepends a learned parent token to the child-memory sequence. Full definitions are provided in Appendix~\ref{app:aggregation}.

\[
\tilde{m}_v \;\triangleq\; \mathrm{Agg}\!\left(M^{\mathrm{ch}}_v\right) \in \mathbb{R}^{d}.
\]

Given \(\tilde{m}_v\), we generate the node memory \(m_v\) by conditioning the backbone on both \(\tilde{m}_v\) and the node text \(x_v\).

Let $E_v \in \mathbb{R}^{n_v \times d}$ denote the token embeddings of $x_v$. We form
\[
\tilde{E}_v \;=\; [\, e_{\mathrm{write}} \,;\; \tilde{m}_v \,;\; E_v \,;\; e_{\mathrm{read}} \,]
\in \mathbb{R}^{(n_v+3)\times d},
\]
run the backbone on $\tilde{E}_v$, and set $m_v$ to the final-layer hidden state at the $e_{\mathrm{read}}$ position. If $x_v$ is empty, we bypass the backbone pass and set $m_v \triangleq \tilde{m}_v$.

\subsection{Hierarchical Memory-Aware Inference}
\label{inference}
Given a question \(q\) and a precomputed semantic hierarchy \(\mathcal{T}=(V,E,r)\), H$^{2}$MT performs coarse-to-fine routing over the cached node memories \(\{m_v \in \mathbb{R}^d : v\in V\}\). The objective is to select relevant branches under a bounded memory budget without reprocessing the raw document.

The intuition aligns with a common pattern in human recall: when people are asked a question about previously memorized material, they typically form a high-level outline first, use it to localize the relevant part of their memory, and only then retrieve the fine-grained details within that region. Without this coarse-to-fine localization, accessing specific details is often slower and more error-prone.

\textbf{Query embedding.} Let \(q=(q_1,\ldots,q_{|q|})\) denote the tokenized question and \(B_{\mathrm{seg}}\) the configured segment size. We retain \(n_q=\min\!\left(|q|,\left\lfloor B_{\mathrm{seg}}/2\right\rfloor\right)\) tokens and form \(\tilde{E}_q=[\,e_{\mathrm{write}};\,E_q[1{:}n_q];\,e_{\mathrm{read}}\,]\). The final-layer hidden state at the read position is used as the query vector \(\mathbf{q}\in\mathbb{R}^{d}\). The cap is an efficiency safeguard for unusually long questions; all evaluated questions fall below it, so no question content is truncated in the reported experiments.

\textbf{Top-down routing with per-parent pruning.} Inference proceeds from the root. For each routed parent \(p\), we score a child \(u\in C(p)\) by \(s(u\mid\mathbf q)=(W_q\mathbf q)^\top(W_km_u)/\sqrt{d_h}\), where \(W_q\) and \(W_k\) are learnable projections. Starting from \(\mathcal S_0=\{r\}\), routing applies
\[
S(p)=\operatorname{TopK}_{u\in C(p)} s(u\mid\mathbf q), \qquad \mathcal S_{\ell+1}=\bigcup_{p\in\mathcal S_\ell}S(p).
\]
For each parent, \(S(p)\) contains up to \(k\) children, or all children when \(|C(p)|<k\). Per-parent selection retains multiple candidate subtrees and avoids collapsing the traversal to a single branch. Traversal terminates at leaves, the maximum depth, or the memory budget.

\textbf{Output generation.} Let \(L_r\) be the final visited depth and \(R=\operatorname{concat}(\mathcal S_0,\ldots,\mathcal S_{L_r})\) the ordered nodes retained across routing levels. We form \(M_{\mathrm{ret}}=\operatorname{stack}\{m_v:v\in R\}\) and prepend these memories to the embedded question for answer generation. Appendix~\ref{app:complexity} gives the worst-case retrieval-size and routing-complexity bounds.


\subsection{Training Objectives}
We optimize \mymodel in two stages: (i) corpus training learns bottom-up memory construction, and (ii) QA fine-tuning optimizes answer generation together with explicit routing and selection objectives. Throughout, each node \(v\) stores a single memory embedding \(m_v \in \mathbb{R}^d\).

\textbf{Corpus training (hierarchical memory construction).}
For each node $v$ with local text $x_v=(x_{v,1},\ldots,x_{v,n_v})$ and memory embedding $m_v$, we implement conditioning on $m_v$ by inserting $m_v$ as an embedding prefix token before the embeddings of $x_v$. We train memory-conditioned modeling using token-level cross-entropy:
\[
\mathcal{L}_{\mathrm{LM}}(v)
\;=\;
-\frac{1}{n_v}\sum_{t=1}^{n_v}\log p_{\theta}\!\left(x_{v,t} \mid x_{v,<t}, m_v\right).
\]
This objective is applied to leaves and internal nodes; for internal nodes, \(m_v\) is constructed from child memories via the bottom-up procedure in Section~\ref{sec:bottom_up}.

\textbf{Memory reconstruction (optional).}
To encourage \(m_v\) to preserve information about \(x_v\), we optionally add a reconstruction loss that decodes \(x_v\) from \(m_v\) under a reconstruction prompt and is token-level cross-entropy.

\textbf{QA fine-tuning (generation).}
Given a question \(q\), retrieved memory context \(M_{\mathrm{ret}}\), and answer tokens \(a=(a_1,\ldots,a_A)\), we minimize:
\[
\mathcal{L}_{\mathrm{gen}}
\;=\;
-\frac{1}{A}\sum_{t=1}^{A}\log p_{\theta}\!\left(a_t \mid q, M_{\mathrm{ret}}, a_{<t}\right).
\]

Because inference uses discrete top-$k$ selection, $\mathcal{L}_{\mathrm{gen}}$ does not backpropagate through the selected node identities into the routing scores. The router is instead trained through $\mathcal{L}_{\mathrm{route}}$ and $\mathcal{L}_{\mathrm{sel}}$, which are computed from a differentiable distribution over the children of each supervised parent.

\textbf{Routing supervision.}
For datasets with annotated evidence, we align evidence spans to leaf-node text by string matching and propagate matched leaf labels to their ancestors. Each routed parent therefore receives a gold child or gold set consisting of branches that contain at least one matched relevant descendant. At each routed parent \(p\), let \(C(p)\) be its children and let \(s_p(u)\) denote the routing score for child \(u\in C(p)\). For gold-path supervision with a single gold child \(g(p)\) at each parent, we use a per-level cross-entropy:
\[
\mathcal{L}_{\mathrm{route}}
\;=\;
-\sum_{p\in \mathcal{P}}
\log
\frac{\exp\!\left(s_p(g(p))/\tau\right)}
{\sum_{u\in C(p)} \exp\!\left(s_p(u)/\tau\right)},
\]
where \(\mathcal{P}\) is the set of routed parents along the supervised path and \(\tau\) is a temperature.

\textbf{Selection supervision (top-\(k\) pruning).}
When pruning selects a set of children (top-\(k\)) rather than a single branch, we supervise the selection using a gold set \(\mathcal{G}(p)\subseteq C(p)\). Let
\[
\pi_p(u) \;\triangleq\; \frac{\exp(s_p(u)/\tau)}{\sum_{u'\in C(p)} \exp(s_p(u')/\tau)}.
\]
We maximize the probability mass assigned to the gold set:
\[
\mathcal{L}_{\mathrm{sel}}
\;=\;
-\sum_{p\in \mathcal{P}}
\log \sum_{u\in \mathcal{G}(p)} \pi_p(u).
\]

\textbf{Overall objective.}
During corpus training:
\[
\mathcal{L}_{\mathrm{corpus}}
\;=\;
\sum_{v\in V}\Big(\mathcal{L}_{\mathrm{LM}}(v) + \lambda_{\mathrm{ae}}\mathcal{L}_{\mathrm{AE}}(v)\Big),
\]
where \(\mathcal{L}_{\mathrm{AE}}\) is the optional reconstruction loss.
During QA fine-tuning:
\[
\mathcal{L}_{\mathrm{QA}}
\;=\;
\mathcal{L}_{\mathrm{gen}}
+ \lambda_r \mathcal{L}_{\mathrm{route}}
+ \lambda_s \mathcal{L}_{\mathrm{sel}}
+ \lambda_{\mathrm{ae}}\!\sum_{v\in \mathcal{V}_{\mathrm{enc}}}\mathcal{L}_{\mathrm{AE}}(v),
\]
where \(\mathcal{V}_{\mathrm{enc}}\subseteq V\) denotes the set of nodes whose memories are trained with reconstruction (e.g., leaves only or all nodes), and \(\lambda_r,\lambda_s,\lambda_{\mathrm{ae}}\) are scalar weights. For NarrativeQA, which does not provide reliable evidence-span supervision, we omit evidence-derived routing and selection supervision. The generation loss still trains answer generation conditioned on the selected memories, but it does not directly supervise the routing scores.

\textbf{Implementation details.}
We optimize the trainable components using AdamW with a learning rate of $10^{-4}$. LoRA uses rank $8$, scaling factor $32$, and dropout $0.1$. QA fine-tuning runs for up to $3000$ steps. Dataset-specific routing budgets, optimization settings, LoRA target
modules, and decoding settings are reported in
Appendix~\ref{app:hyper}.


\section{Experiments}
\label{sec:experiments}

We organize the evaluation around four research questions:

\begin{itemize}
    \item \textbf{RQ1:} Does H\textsuperscript{2}MT provide a favorable
    quality--efficiency trade-off on intrinsically structured documents?
    \item \textbf{RQ2:} What is the practical contribution of hierarchical routing?
    \item \textbf{RQ3:} How do the per-parent routing budget $k$ and intermediate-node aggregation affect quality and efficiency?
    \item \textbf{RQ4:} When does offline hierarchy and memory construction
    amortize across repeated queries?
\end{itemize}

\subsection{Experimental Setup}
\label{sec:experimental-setup}

\paragraph{Datasets and evaluation regimes.}
We evaluate H\textsuperscript{2}MT under three document-structure regimes. QASPER contains questions grounded in full scientific papers, whose section and subsection organization provides an explicit document
hierarchy. For OpenROAD, we use ORD-QA \citep{pu2024customized}, a question-answering benchmark grounded in the documentation of the open-source OpenROAD electronic-design-automation toolchain.

NarrativeQA provides an additional long document evaluation. We represent each narrative as document $\rightarrow$ chapter or scene $\rightarrow$ paragraph leaves using the structural boundaries produced by our preprocessing pipeline. NarrativeQA evaluates long-document generation without reliable annotated evidence spans.

HotpotQA serves as a weak-structure stress test. Each example contains multiple Wikipedia articles, but the dataset does not provide an intrinsic hierarchy connecting those articles. We therefore use HotpotQA to evaluate the operating boundary of hierarchy-aware routing,
rather than as evidence for the main structured-document claim.

We additionally report results on a private VLSI layout-verification documentation collection as a technical domain case study.

\paragraph{Hierarchy construction.}
Hierarchy construction depends on the structure available in each dataset. When explicit document organization is available, we preserve that organization rather than inducing a new tree. For QASPER, the hierarchy follows paper section or subsection, paragraph or chunk. For OpenROAD, headings define internal nodes, and text scoped under each heading is attached to the corresponding node or descendant leaves. For NarrativeQA, the hierarchy follows document chapter or scene, paragraph leaves.

For the HotpotQA setting, each supplied article is represented independently as an article-level root with sentence descendants. This shallow organization groups evidence within individual articles but does not encode semantic relations across articles. We therefore treat it as a weak-structure baseline. We separately evaluate a hierarchy induced by recursive GMM clustering over node memories, denoted H\textsuperscript{2}MT--GMM.

Table~\ref{tab:hierarchy-routing} summarizes the hierarchy and routing configurations. Searchable-node and selected-memory counts are configuration-specific and should be interpreted within the stated paper, corpus, book, article, or document scope.

\begin{table}[t]
\centering
\scriptsize
\setlength{\tabcolsep}{5pt}
\renewcommand{\arraystretch}{1.12}
\caption{Hierarchy and routing statistics. Node and selected-memory
counts are averages within the routing unit shown for each dataset.}
\label{tab:hierarchy-routing}
\begin{tabular}{llrrr}
\toprule
Dataset
& Routing unit
& Nodes/unit
& Per-parent \(k\)
& Selected/unit \\
\midrule
QASPER
& One paper
& 15.6
& 8
& 8.4 \\

OpenROAD
& Entire corpus
& 447
& 16
& 26.4 \\

NarrativeQA
& One book
& 990.5
& 16
& 254.0 \\

HotpotQA
& One article
& 5.11
& 16
& 3.3 \\
\bottomrule
\end{tabular}
\end{table}

A HotpotQA example contains multiple articles. Because
each article is represented and routed independently, the HotpotQA
row reports per-article averages.

\paragraph{External baselines.}
We compare H\textsuperscript{2}MT with memory-based, prompt-compression, and text-retrieval methods. Memory-based baselines include HMT \citep{he-etal-2025-hmt}, ICAE
\citep{ge2024incontext}, MemoryLLM \citep{pmlr-v235-wang24s}, and UltraGist / Activation Beacon~\cite{zhang2024activationbeacon}. Input-side compression baselines
include LongLLMLingua \citep{jiang-etal-2024-longllmlingua} and AutoCompressor
\citep{chevalier-etal-2023-adapting}. We additionally compare against RAPTOR \citep{sarthi2024raptor}, a RAG-based method that retrieves textual nodes from a recursively constructed summary tree.

Because these systems use different memory representations, checkpoints, and supported backbones, we use them to characterize the overall quality--efficiency frontier rather than to isolate the effect of hierarchy.

\paragraph{Evaluation metrics.} For NarrativeQA, HotpotQA, QASPER, and OpenROAD, we report dataset-level ROUGE-L in Table~\ref{tab:main_rouge} and average TTFT and peak GPU memory in Table~\ref{tab:avg_efficiency}. Full dataset-level efficiency results are reported in Appendix~\ref{app:full-efficiency}. We additionally report F1 on NarrativeQA, QASPER, and HotpotQA for LLaMA-2-7B in Table~\ref{tab:llama2_7b_nqa_qasper_hpqa}. All baseline rows in this table were reproduced under our common evaluation protocol; Appendix~\ref{app:baseline-configs} reports the checkpoint or code source, data-specific instruction prompt template, context and compression budgets, decoding settings, and deviations from the original implementations. Results on the private VLSI layout-verification collection are reported in Appendix~\ref{app:lvs}.

\paragraph{Evaluation protocol.}
Unless otherwise stated, decoding is greedy. TTFT is measured from request submission to the first decoded token, total generation time is measured to the final decoded token, and peak GPU memory is the maximum allocated memory over the run. Dataset-specific routing budgets and training details are reported in Appendix~\ref{app:hyper}.

\subsection{RQ1: Does \mymodel Improve the Quality--Efficiency Trade-off on Structured Documents?}
\label{sec:rq1}

\paragraph{Main quality--efficiency comparison.}

We first evaluate answer quality, time to first token, and peak GPU memory. QASPER and OpenROAD constitute the primary structured-document evaluation, while HotpotQA serves as a weak-structure stress test. Table~\ref{tab:main_rouge} reports dataset-level quality, while Table~\ref{tab:avg_efficiency} summarizes average TTFT and peak GPU memory.

\begin{table*}[t]
\centering
\footnotesize
\setlength{\tabcolsep}{0pt}
\renewcommand{\arraystretch}{0.95}
\setlength{\aboverulesep}{0pt}
\setlength{\belowrulesep}{0pt}

\caption{ROUGE-L score (\%, higher is better) on HotpotQA, NarrativeQA, QASPER, and OpenROAD. Best \emph{per backbone and dataset} is in bold. Avg is the unweighted mean over the four datasets; best Avg \emph{per backbone} is in bold.}
\label{tab:main_rouge}

\begin{tabular}{@{}l@{\hspace{6pt}}l@{\hspace{10pt}}r@{\hspace{10pt}}r@{\hspace{10pt}}r@{\hspace{10pt}}r@{\hspace{10pt}}>{\columncolor{avgblue}}r@{}}
\toprule
Model & Method & HotpotQA & NarrativeQA & QASPER & OpenROAD & Avg \\
\midrule

\multirow{3}{*}{Qwen2.5-14B}
& Backbone & 19.00 & 9.96 & 10.13 & 27.37 & 16.62 \\
& HMT      & \textbf{42.40} & 29.17 & 29.70 & 32.00 & 33.32 \\
& H$^2$MT  & 30.37 & \textbf{41.67} & \textbf{38.80} & \textbf{32.60} & \textbf{35.86} \\
\cmidrule(lr){1-7}

\multirow{3}{*}{Qwen2.5-7B}
& Backbone & 19.60 & 9.46 & 12.08 & 27.44 & 17.15 \\
& HMT      & \textbf{30.06} & 14.42 & 14.51 & 26.09 & 21.27 \\
& H$^2$MT  & 28.89 & \textbf{29.26} & \textbf{34.15} & \textbf{30.65} & \textbf{30.74} \\
\cmidrule(lr){1-7}

\multirow{3}{*}{Qwen3-4B}
& Backbone & 18.21 & 12.15 & 10.92 & 25.13 & 16.60 \\
& HMT      & 28.21 & 10.01 & 15.74 & \textbf{33.41} & 21.84 \\
& H$^2$MT  & \textbf{30.37} & \textbf{45.00} & \textbf{31.29} & 32.95 & \textbf{34.90} \\
\cmidrule(lr){1-7}

\multirow{3}{*}{Llama-3.1-8B}
& Backbone & 19.99 & 22.84 & 17.99 & 27.30 & 22.03 \\
& HMT      & \textbf{30.54} & 21.02 & 27.00 & \textbf{29.40} & 26.99 \\
& H$^2$MT  & 25.37 & \textbf{42.83} & \textbf{28.72} & 28.80 & \textbf{31.43} \\
\cmidrule(lr){1-7}

\multirow{3}{*}{SmolLM-135M}
& Backbone & 3.76 & 1.04 & 11.54 & 18.47 & 8.70 \\
& HMT      & \textbf{27.44} & 16.44 & 20.30 & 22.76 & 21.74 \\
& H$^2$MT  & 23.80 & \textbf{28.57} & \textbf{27.30} & \textbf{24.40} & \textbf{26.02} \\
\bottomrule
\end{tabular}
\end{table*}

\begin{table}[t]
\centering
\footnotesize
\setlength{\tabcolsep}{4.5pt}
\renewcommand{\arraystretch}{1.12}

\caption{Average inference efficiency across HotpotQA, NarrativeQA,
QASPER, and OpenROAD. TTFT is in milliseconds and peak GPU memory
is in GB; lower is better. Best per backbone is in bold. Full
dataset-level results are reported in
Appendix~\ref{app:full-efficiency}.}
\label{tab:avg_efficiency}

\begin{tabular}{@{}llrr@{}}
\toprule
Model & Method & Avg. TTFT & Avg. peak mem. \\
& & (ms) & (GB) \\
\midrule

\multirow{2}{*}{Qwen2.5-14B}
& HMT     & 2897.49 & 56.85 \\
& H$^2$MT & \textbf{468.92} & \textbf{42.90} \\
\cmidrule(lr){1-4}

\multirow{2}{*}{Qwen2.5-7B}
& HMT     & 6398.96 & 30.46 \\
& H$^2$MT & \textbf{276.89} & \textbf{23.17} \\
\cmidrule(lr){1-4}

\multirow{2}{*}{Qwen3-4B}
& HMT     & 5401.29 & 14.91 \\
& H$^2$MT & \textbf{302.31} & \textbf{14.60} \\
\cmidrule(lr){1-4}

\multirow{2}{*}{Llama-3.1-8B}
& HMT     & 2154.94 & 31.17 \\
& H$^2$MT & \textbf{369.77} & \textbf{24.33} \\
\cmidrule(lr){1-4}

\multirow{2}{*}{SmolLM-135M}
& HMT     & 577.50 & 0.85 \\
& H$^2$MT & \textbf{477.41} & \textbf{0.65} \\
\bottomrule
\end{tabular}
\end{table}

\begin{table}[t]
\centering
\small
\setlength{\tabcolsep}{4pt}
\renewcommand{\arraystretch}{1.12}
\caption{Performance (F1) on NarrativeQA (NQA), QASPER, and HotpotQA for Llama-2-7B under our common evaluation harness. Best per column is in bold.}
\label{tab:llama2_7b_nqa_qasper_hpqa}
\begin{tabular}{lccc}
\toprule
Method & NQA & QASPER & HotpotQA \\
\midrule
Llama-2-7B         & 18.70 & 19.20 & 25.40 \\
AutoCompressors    & 11.89 & 11.64 & 16.32 \\
LongLLMLingua      & 16.54 & 21.10 & 20.50 \\
ICAE               & 17.17 & 23.35 & 32.16 \\
MemoryLLM-16k      & 13.40 & 19.57 & 15.58 \\
MemoryLLM-all-BM25 & 15.60 & 20.30 & 15.23 \\
UltraGist & 16.96 & 34.49 & \textbf{34.07} \\
\midrule
H$^{2}$MT (ours)   & \textbf{24.90} & \textbf{36.67} & 23.80 \\
\bottomrule
\end{tabular}
\end{table}

\paragraph{Quality and TTFT}
Across the  backbones with answer-quality results, H$^{2}$MT obtains the highest average ROUGE-L for every backbone. The gains concentrate on NarrativeQA and QASPER, whereas HMT remains stronger on HotpotQA for four of the five backbones, consistent with the weaker hierarchy available for multi-document inputs. Across all five backbones, H$^{2}$MT achieves lower average TTFT than HMT
after averaging over the four datasets. The improvement is not universal
at the dataset level: on NarrativeQA with SmolLM-135M, H$^{2}$MT is
slower than HMT, with TTFT of 1127.05 versus 816.82\,ms.
Appendix~\ref{app:full-efficiency} reports the complete results.

\paragraph{Peak GPU memory.} We report peak GPU memory (GB), defined as the maximum GPU memory during inference. This measurement includes model weights, activations, the KV cache, temporary buffers, and any other tensors allocated during inference. Peak memory captures the worst-case footprint, so it directly answers whether a configuration fits on a given GPU. Against hierarchical retrieval, H$^{2}$MT uses less peak GPU memory on QASPER: 10.46 GB versus 15.97 GB with Qwen3-4B and 13.46 GB versus 26.57 GB with Llama-2-7B. On OpenROAD, H$^{2}$MT uses slightly more memory than RAPTOR, 17.10 versus 16.07 GB, while on HotpotQA the two remain within two percent of each other (Table~\ref{tab:raptor}). Against memory-based baselines, H$^{2}$MT lowers average peak memory for every backbone, although the dataset-level reduction is not universal; complete results and isolated reversals are reported in Appendix~\ref{app:full-efficiency}.

\paragraph{Relationship between TTFT and peak GPU memory.}
TTFT and peak GPU memory capture different bottlenecks and do not necessarily move together. TTFT is dominated by prefill latency up to the first generated token, while peak memory reflects the maximum footprint over the entire run, often dominated by KV-cache growth during decoding or by temporary buffers. As a result, reducing TTFT by shortening the effective prefill can still increase peak memory when routing retrieves more nodes (larger $k$) or when deeper traversal increases the number of memory tokens inserted.

\paragraph{Technical-domain case study.}
On the private LVS technical-document corpus, H$^{2}$MT reduces
held-out perplexity from 11.93 with HMT to 2.62, corresponding to
a 78.1\% reduction, or \(4.56\times\) lower perplexity. Because LVS is not publicly available,
we treat this result as supporting domain-specific evidence rather
than part of the main public-benchmark comparison; Appendix~D
provides the dataset construction and complete results.

\subsection{Comparison with Selective-Attention and KV-Cache Systems}
\label{sec:kvcache}

Selective-attention and KV-cache systems manage token-level states, whereas H\textsuperscript{2}MT changes the representation supplied to the reader. RetroInfer and ArkVale reduce the storage or access cost of token-level KV states, while H\textsuperscript{2}MT routes over precomputed node memories and avoids reprocessing the complete raw document at query time. The approaches are therefore complementary, but direct latency comparisons require matched hardware, backbone, context length, and decoding settings.

We evaluated RetroInfer and ArkVale on QASPER with
Llama-2-7B-Chat using 200 LongBench examples and greedy decoding.
Table~\ref{tab:kv-comparison} reports the results. Because the
systems were evaluated on different hardware, the TTFT values
characterize their observed implementations but do not constitute
a controlled latency comparison.

\begin{table}[t]
\centering
\small
\setlength{\tabcolsep}{7pt}
\renewcommand{\arraystretch}{1.05}
\caption{Auxiliary comparison with selective-attention and KV-cache
systems on QASPER. All methods use Llama-2-7B-family backbones and
are evaluated on 200 LongBench examples with greedy decoding.
RetroInfer and ArkVale use an NVIDIA A100-40GB, whereas
H$^{2}$MT uses AMD hardware; TTFT is therefore not directly
comparable across systems.}
\label{tab:kv-comparison}
\begin{tabular}{lrr}
\toprule
Method & ROUGE-L & TTFT (ms) \\
\midrule
RetroInfer & 22.8 & 2353 \\
ArkVale & 20.5 & 470 \\
H$^{2}$MT & 36.33 & 275.5 \\
\bottomrule
\end{tabular}
\end{table}

\subsection{RQ2: What Is the Contribution of Hierarchical Routing?}
\label{sec:rq2}

\paragraph{Routed selection versus loading the complete hierarchy.}
On QASPER with Qwen3-4B and \(k=8\), routed selection retains
8.4 node memories on average and obtains 31.29 ROUGE-L, compared
with 30.60 ROUGE-L when all 15.6 node memories are supplied to
the reader. Loading all nodes can introduce parent--child redundancy,
because internal-node memories aggregate information also represented
by their descendants. This result should not be interpreted as showing that pruning
universally improves answer quality. Its direct benefit is to bound
the reader-side memory context, while its quality effect depends on
whether the excluded nodes contain useful evidence or primarily
introduce redundant and distracting information.

\paragraph{Relation to flat memory retrieval.}
A leaf-only version of H\textsuperscript{2}MT without intermediate-node aggregation or top-down traversal would reduce the hierarchy to a flat collection of segment memories. HMT provides an informative reference for this regime, but it also differs in recurrent memory construction, training, and retrieval. We therefore treat HMT as an external flat-memory reference rather than as a strictly controlled hierarchy ablation.

\begin{table*}[t]
\caption{Retrieval-augmented generation comparison. ROUGE-L is in \%;
TTFT and total generation time are in ms; peak GPU memory is in GB.
Tree build is a one-time offline construction cost, reported per paper, per
corpus, or per document according to the search scope of each
configuration. The better of the two methods in each configuration is
in bold.}
\label{tab:raptor}
\centering
\scriptsize
\setlength{\tabcolsep}{5pt}
\renewcommand{\arraystretch}{1.20}
\begin{tabular}{lllrrrrr}
\toprule
Dataset & Backbone & Method & ROUGE-L & Tree build & TTFT & Total gen. & Peak GPU \\
\midrule
\multirow{4}{*}{QASPER}
 & \multirow{2}{*}{Qwen3-4B}
   & SBERT+RAPTOR & 16.99 & 762\,s/paper & \textbf{52.3}
   & \textbf{1490} & 15.97 \\
 & & H$^{2}$MT & \textbf{31.29} & \textbf{7.50\,s/paper}
   & 306.99 & 2584.8 & \textbf{10.46} \\
\cmidrule(l){2-8}
 & \multirow{2}{*}{Llama-2-7B}
   & SBERT+RAPTOR & 25.44 & 762\,s/paper & \textbf{60.2}
   & \textbf{685} & 26.57 \\
 & & H$^{2}$MT & \textbf{36.33} & \textbf{3.84\,s/paper}
   & 275.5
   & 2404.7
   & \textbf{13.46} \\
\midrule
\multirow{2}{*}{OpenROAD} & \multirow{2}{*}{Qwen3-4B}
   & SBERT+RAPTOR & 24.03 & 11709\,s/corpus & \textbf{66.8}
   & 5124 & \textbf{16.07} \\
 & & H$^{2}$MT & \textbf{32.95} & \textbf{869\,s/corpus}
   & 321.18 & \textbf{1129.6} & 17.10 \\
\midrule
\multirow{2}{*}{HotpotQA} & \multirow{2}{*}{Qwen3-4B}
   & SBERT+RAPTOR & \textbf{56.39} & 26.78\,s/doc & \textbf{82.7}
   & 11864.20 & 16.08 \\
 & & H$^{2}$MT (GMM) & 46.64 & \textbf{0.69\,s/doc}
   & 219.72 & \textbf{1661.15} & \textbf{15.79} \\
\bottomrule
\end{tabular}
\vspace{2pt}
\end{table*}

\subsection{RQ3: How Do the Routing Budget and Intermediate-Node Aggregation Affect Quality and Efficiency?}
\label{sec:rq3}

\begin{figure}[t]
  \centering
  \includegraphics[width=\columnwidth]{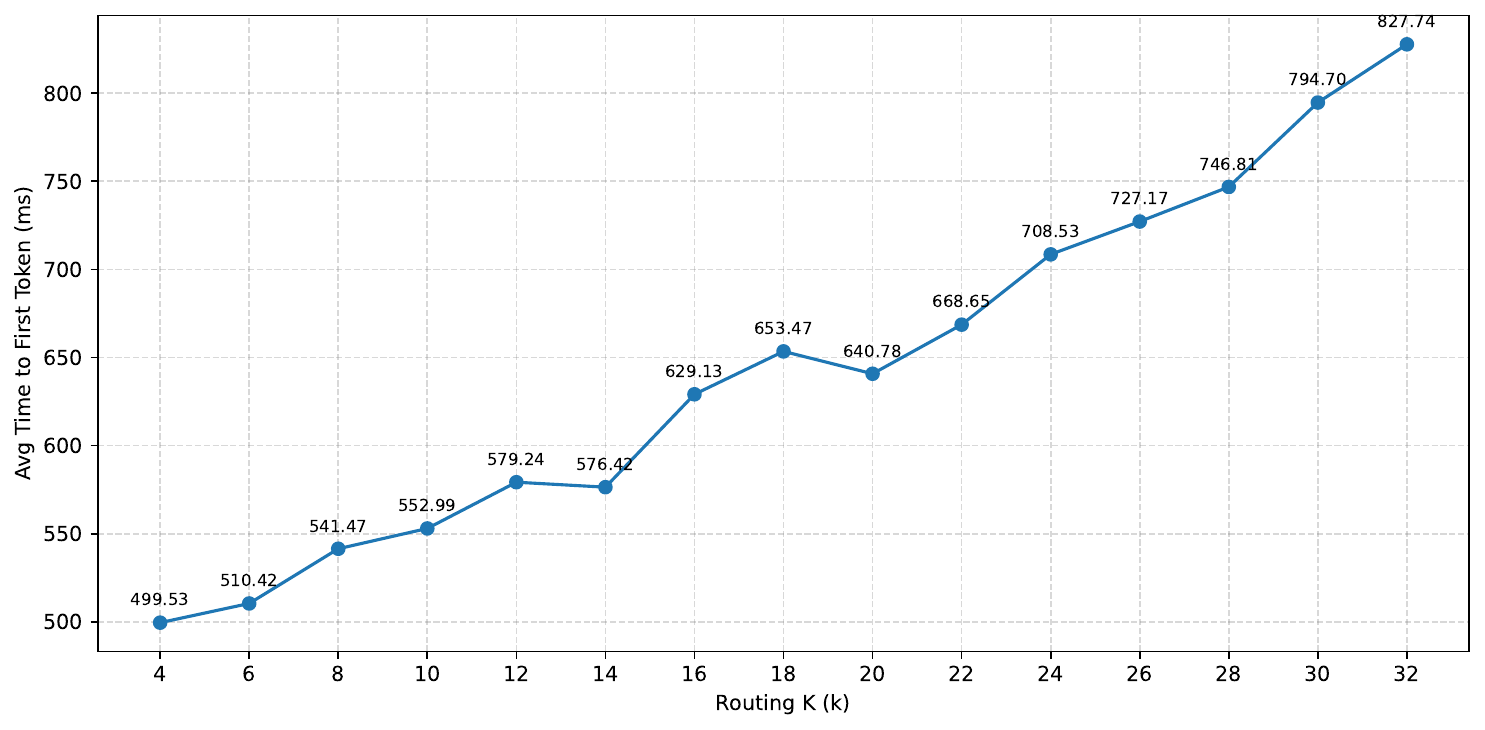}
  \caption{Average TTFT as a function of the per-parent routing budget \(k\) on NarrativeQA using Qwen2.5-14B-Instruct. TTFT increases from \(499.5\) ms at \(k=4\) to \(827.7\) ms at \(k=32\).}
  \label{fig:k_sweep}
\end{figure}

\paragraph{Effect of top-\(k\) selection on efficiency.}
Figure~\ref{fig:k_sweep} shows that increasing \(k\) from 4 to 32 raises average TTFT from \(499.5\) to \(827.7\) ms on NarrativeQA using Qwen2.5-14B-Instruct, because more node memories are selected before generation. In the same configuration, peak GPU memory varies by less than \(0.05\) GB, under \(0.2\%\) of the total footprint. Thus, \(k\) primarily controls latency rather than peak memory in this setting. Appendix~\ref{app:routing-memory} reports the complete peak-memory sweep.

\begin{figure}[t]
  \centering
  \includegraphics[width=0.9\columnwidth]{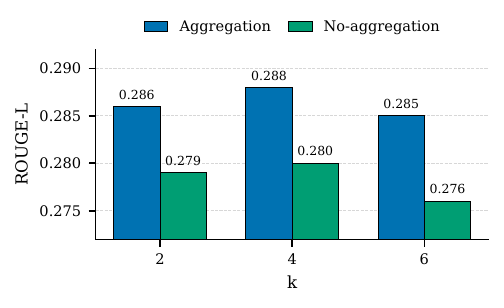}
    \caption{Controlled routing-budget ablation on OpenROAD using LLaMA-3.1-8B. Bars report ROUGE-L with and without intermediate-node aggregation for \(k\in\{2,4,6\}\).}
  \label{fig:rougel-vs-k}
\end{figure}

\paragraph{Effect of $k$ and intermediate-node aggregation on quality.}
Figure~\ref{fig:rougel-vs-k} analyzes how the per-parent routing budget $k$ interacts with intermediate-node aggregation. We observe a sweet spot: increasing $k$ from small values improves ROUGE-L by covering more relevant branches, while overly large $k$ admits weakly related nodes that dilute the selected context and reduce answer focus. Aggregation reduces this sensitivity by making each internal node a compact representation of its descendant content during offline construction; in this experiment, we use cross-attention aggregation. Without aggregation, internal nodes are less informative and routing becomes noisier, so the retrieved set accumulates more redundant or irrelevant content as $k$ grows. Overall, moderate $k$ performs best, and aggregation helps control distractors when widening the routing budget.

\subsection{RQ4: When Does Offline Construction Amortize across Repeated Queries?}
\label{sec:rq4}
H$^{2}$MT separates one-time hierarchy and memory construction from per-query routing and generation. As shown in Table~\ref{tab:raptor}, H$^{2}$MT has both lower construction cost and lower total generation time on OpenROAD and HotpotQA, so it is cheaper from the first query. On QASPER, H$^{2}$MT has substantially lower construction cost but higher per-query generation time; it remains cheaper up to approximately 689 queries per paper with Qwen3-4B and 441 with Llama-2-7B. Appendix~\ref{app:amortization} provides the complete crossover derivation and LVS analysis.

\subsection{Induced-Hierarchy Extension on HotpotQA}
\label{sec:gmm-hotpot}

As an exploratory weak-structure extension, we construct a separate hierarchy for each HotpotQA document through recursive GMM clustering over chunk memories. This improves H$^{2}$MT from 30.37 to 46.64 ROUGE-L, but remains below SBERT+RAPTOR at 56.39, confirming that induced latent hierarchies do not eliminate the weak-structure limitation. Appendix~\ref{app:gmm-hotpot} provides the construction details.

\subsection{Retrieval-Augmented Generation Baseline}
\label{sec:raptor}

We additionally compare against RAPTOR~\citep{sarthi2024raptor}, which constructs a hierarchical text index and supplies retrieved chunks and summaries to the generator. Unlike RAPTOR, H$^{2}$MT routes over fixed-size latent node memories and does not concatenate retrieved document text at inference.

The comparison is regime-dependent. On the structured QASPER and OpenROAD settings, H$^{2}$MT exceeds RAPTOR by 8.9--14.3 ROUGE-L points and has substantially lower offline construction cost. On weakly structured HotpotQA, RAPTOR remains stronger in quality, 56.39 versus 46.64, consistent with its use of explicit retrieved text and the sensitivity of H$^{2}$MT to hierarchy-induction and compression errors. RAPTOR reaches the first token sooner in every measured configuration, whereas H$^{2}$MT has lower total generation time on OpenROAD and HotpotQA.

RAPTOR encodes the query with a lightweight sentence encoder and performs one vector search before reader prefill. H$^{2}$MT instead encodes the question with the full backbone and performs iterative per-level routing, so its TTFT includes query encoding, candidate loading, and repeated routing operations. This overhead grows with the explored routing frontier, consistent with Figure~\ref{fig:k_sweep} and the selected-memory statistics in Table~\ref{tab:hierarchy-routing}. These results support H$^{2}$MT primarily for repeated QA over documents with reliable intrinsic structure, rather than as a uniform replacement for raw-text retrieval. Appendix~\ref{app:rag-analysis} provides additional analysis of the representation and TTFT differences.

\paragraph{Combining hierarchical retrieval and prompt compression.}
A complementary text-based baseline could apply prompt compression,
such as LongLLMLingua, to the text selected by RAPTOR. We did not
evaluate this combined pipeline. It would retain RAPTOR's one-time
tree-construction cost and add a query-time compression pass over
the retrieved text, so its quality--latency trade-off cannot be
inferred directly from either component alone. Evaluating this
combination is an important direction for future work.

\section{Conclusion}
We presented H$^{2}$MT, a hierarchy-aware latent-memory framework for repeated question answering over long documents. H$^{2}$MT constructs node memories offline through bottom-up aggregation and performs online coarse-to-fine routing with per-parent pruning to select a bounded query-specific memory context. The method preserves native document structure without modifying the backbone attention architecture. Across the evaluated backbones, H$^{2}$MT improves average ROUGE-L over HMT and substantially reduces HMT's time to first token, with the clearest quality gains on NarrativeQA and QASPER. Peak GPU memory is often lower than HMT's, although the reduction is not universal. Against RAPTOR, H$^{2}$MT achieves higher answer quality and substantially lower offline construction cost on the structured QASPER and OpenROAD settings, while RAPTOR reaches the first token sooner and remains stronger on the weakly structured HotpotQA setting.

\paragraph{Limitations.}
H$^{2}$MT relies on a meaningful document hierarchy. For weakly structured corpora such as HotpotQA, induced hierarchies are less reliable than native document structure, and clustering errors can propagate to memory construction and routing. Moreover, compressing a node's descendant content into a single memory vector may attenuate fine-grained evidence. Because routing applies hard early pruning, evidence discarded at a higher level cannot be recovered during later traversal.

\paragraph{Future work.} Future work will explore richer structures beyond trees, including
cross-reference and shared-entity links, together with routing objectives
that better preserve rare but critical evidence under strict memory
budgets.


\bibliographystyle{ACM-Reference-Format}
\bibliography{references}

\newpage
\appendix
\onecolumn

\section{Core Notation}
\label{app:notation}

\begin{table}[H]
\centering
\scriptsize
\setlength{\tabcolsep}{5pt}
\renewcommand{\arraystretch}{1}
\caption{Notation summary.}
\label{tab:notation}
\begin{tabularx}{\columnwidth}{@{}lX@{}}
\toprule
Symbol & Description \\
\midrule
$\mathcal{T}=(V,E,r)$ & Semantic hierarchy tree with nodes $V$, edges $E$, and root $r$. \\
$p(\cdot)$, $C(v)$, $c_v$ & Parent map; children set; $c_v=|C(v)|$. \\
$\mathcal{T}_v$ & Subtree rooted at $v$. \\
$x_v$, $n_v$ & Local text at node $v$; token count. \\
$d$ & Backbone hidden dimension. \\
$e_{\mathrm{write}}, e_{\mathrm{read}}$ & Learnable write/read embeddings. \\
$m_v$ & Node memory embedding (one vector per node). \\
$M^{\mathrm{ch}}_v$, $\tilde m_v$ & Child-memory stack; $\tilde m_v=\mathrm{Agg}(\cdot)$ over child (and optional parent) tokens. \\
$q$, $\mathbf{q}$, $n_q$ & Question text; query vector; number of question tokens used to form $\mathbf{q}$. \\
$s(u\mid \mathbf{q})$ & Routing score between query vector $\mathbf{q}$ and node $u$. \\
$W_q,W_k$, $d_h$ & Projections and head dim in the routing score. \\
$k$, $\mathcal{S}_\ell$, $S(p)$ & Per-parent top-$k$; candidates at depth $\ell$; selected children. \\
$\mathcal{R}$, $M_{\mathrm{ret}}$ & Ordered retrieval sequence; stacked retrieved memories. \\
\bottomrule
\end{tabularx}
\end{table}

\section{Full Dataset-Level Efficiency Results}
\label{app:full-efficiency}
\begin{table}[!ht]
\centering
\footnotesize
\setlength{\tabcolsep}{0pt}
\renewcommand{\arraystretch}{0.95}
\setlength{\aboverulesep}{0pt}
\setlength{\belowrulesep}{0pt}

\caption{Average TTFT (ms, lower is better) on HotpotQA, NarrativeQA, QASPER, and OpenROAD. Avg is the unweighted mean over the four datasets; best (lowest) Avg \emph{per backbone} is in bold.}
\label{tab:four_datasets_ttft_rows_h2mt_hmt_backbone}

\begin{tabular}{@{}l@{\hspace{6pt}}l@{\hspace{10pt}}r@{\hspace{10pt}}r@{\hspace{10pt}}r@{\hspace{10pt}}r@{\hspace{10pt}}>{\columncolor{avgblue}}r@{}}
\toprule
Model & Method & HotpotQA & NarrativeQA & QASPER & OpenROAD & Avg \\
\midrule

\multirow{2}{*}{Qwen2.5-14B}
& HMT      & 1959.66 & 3537.69 & 3841.90 & 2250.71 & 2897.49 \\
& H$^2$MT  & \textbf{345.11}  & \textbf{643.57}  & \textbf{364.88}  & \textbf{522.13}  & \textbf{468.92} \\
\cmidrule(lr){1-7}

\multirow{2}{*}{Qwen2.5-7B}
& HMT      & 893.38  & 3056.91 & 3056.91 & 18588.63 & 6398.96 \\
& H$^2$MT  & \textbf{204.00}  & \textbf{361.38}  & \textbf{233.20}  & \textbf{308.98}  & \textbf{276.89} \\
\cmidrule(lr){1-7}

\multirow{2}{*}{Qwen3-4B}
& HMT      & 614.77  & 2405.94 & 1696.10 & 16888.34 & 5401.29 \\
& H$^2$MT  & \textbf{225.94}  & \textbf{355.12}  & \textbf{306.99}  & \textbf{321.18}  & \textbf{302.31} \\
\cmidrule(lr){1-7}

\multirow{2}{*}{Llama-3.1-8B}
& HMT      & 992.73  & 3714.65 & 2400.30 & 1512.08 & 2154.94 \\
& H$^2$MT  & \textbf{498.46}  & \textbf{364.12}  & \textbf{267.93}  & \textbf{348.56}  & \textbf{369.77} \\
\cmidrule(lr){1-7}

\multirow{2}{*}{SmolLM-135M}
& HMT      & 259.75  & \textbf{816.82}  & 666.50  & 566.93 & 577.50 \\
& H$^2$MT  & \textbf{221.70}  & 1127.05 & \textbf{289.00}  & \textbf{271.87}  & \textbf{477.41} \\
\bottomrule
\end{tabular}
\end{table}

\begin{table}[!ht]
\centering
\footnotesize
\setlength{\tabcolsep}{0pt}
\renewcommand{\arraystretch}{0.95}
\setlength{\aboverulesep}{0pt}
\setlength{\belowrulesep}{0pt}

\caption{Average peak GPU memory (GB, lower is better) during inference. Avg is the unweighted mean over the four datasets; best (lowest) Avg \emph{per backbone} is in bold. Peak memory is measured over the full run and may not correlate with TTFT.}
\label{tab:four_datasets_gpu_rows_h2mt_hmt_backbone}

\begin{tabular}{@{}l@{\hspace{6pt}}l@{\hspace{10pt}}r@{\hspace{10pt}}r@{\hspace{10pt}}r@{\hspace{10pt}}r@{\hspace{10pt}}>{\columncolor{avgblue}}r@{}}
\toprule
Model & Method & HotpotQA & NarrativeQA & QASPER & OpenROAD & Avg \\
\midrule

\multirow{2}{*}{Qwen2.5-14B}
& HMT      & 57.09 & 57.12 & 54.21 & 58.99 & 56.85 \\
& H$^2$MT  & \textbf{55.68} & \textbf{31.81} & \textbf{28.27} & \textbf{55.85} & \textbf{42.90} \\
\cmidrule(lr){1-7}

\multirow{2}{*}{Qwen2.5-7B}
& HMT      & 30.24 & 30.31 & 28.52 & 32.77 & 30.46 \\
& H$^2$MT  & \textbf{28.91} & \textbf{16.86} & \textbf{17.46} & \textbf{29.43} & \textbf{23.17} \\
\cmidrule(lr){1-7}

\multirow{2}{*}{Qwen3-4B}
& HMT      & \textbf{12.86} & \textbf{13.23} & 15.24 & 18.31 & 14.91 \\
& H$^2$MT  & 15.40 & 15.42 & \textbf{10.46} & \textbf{17.10} & \textbf{14.60} \\
\cmidrule(lr){1-7}

\multirow{2}{*}{Llama-3.1-8B}
& HMT      & 31.45 & 31.52 & 29.54 & 32.15 & 31.17 \\
& H$^2$MT  & \textbf{30.41} & \textbf{18.14} & \textbf{18.24} & \textbf{30.53} & \textbf{24.33} \\
\cmidrule(lr){1-7}

\multirow{2}{*}{SmolLM-135M}
& HMT      & 0.95 & 0.97 & \textbf{0.53} & 0.96 & 0.85 \\
& H$^2$MT  & \textbf{0.66} & \textbf{0.58} & 0.71 & \textbf{0.66} & \textbf{0.65} \\
\bottomrule
\end{tabular}
\end{table}

\section{Routing Ablations}
\label{app:routing-ablation}

\subsection{Routed Selection versus All Nodes}

\begin{table}[H]
\centering
\scriptsize
\setlength{\tabcolsep}{4pt}
\renewcommand{\arraystretch}{1.0}
\caption{Routed selection versus loading the complete hierarchy on QASPER using Qwen3-4B.}
\label{tab:routed-vs-all}
\begin{tabular}{lrr}
\toprule
Condition & ROUGE-L & Avg. memories \\
\midrule
Routed top-\(k\) & 31.29 & 8.4 \\
All nodes & 30.6 & 15.6 \\
\bottomrule
\end{tabular}
\end{table}

\subsection{Peak-Memory Sensitivity to Routing Budget}
\label{app:routing-memory}

We measure peak GPU memory while varying the per-parent routing budget \(k\) from 4 to 32 on NarrativeQA using Qwen2.5-14B-Instruct. Peak memory varies by less than \(0.05\) GB across the sweep, under \(0.2\%\) of the approximately \(31\) GB total footprint. In contrast, average TTFT increases from \(499.5\) to \(827.7\) ms. This result indicates that \(k\) primarily controls prefill latency rather than peak memory in this configuration. It should not be interpreted as establishing that selected memories have no memory cost, since peak allocation can also depend on temporary workspaces, generated sequence length, and allocator behavior.

\begin{figure}[H]
\centering
\includegraphics[width=\columnwidth]{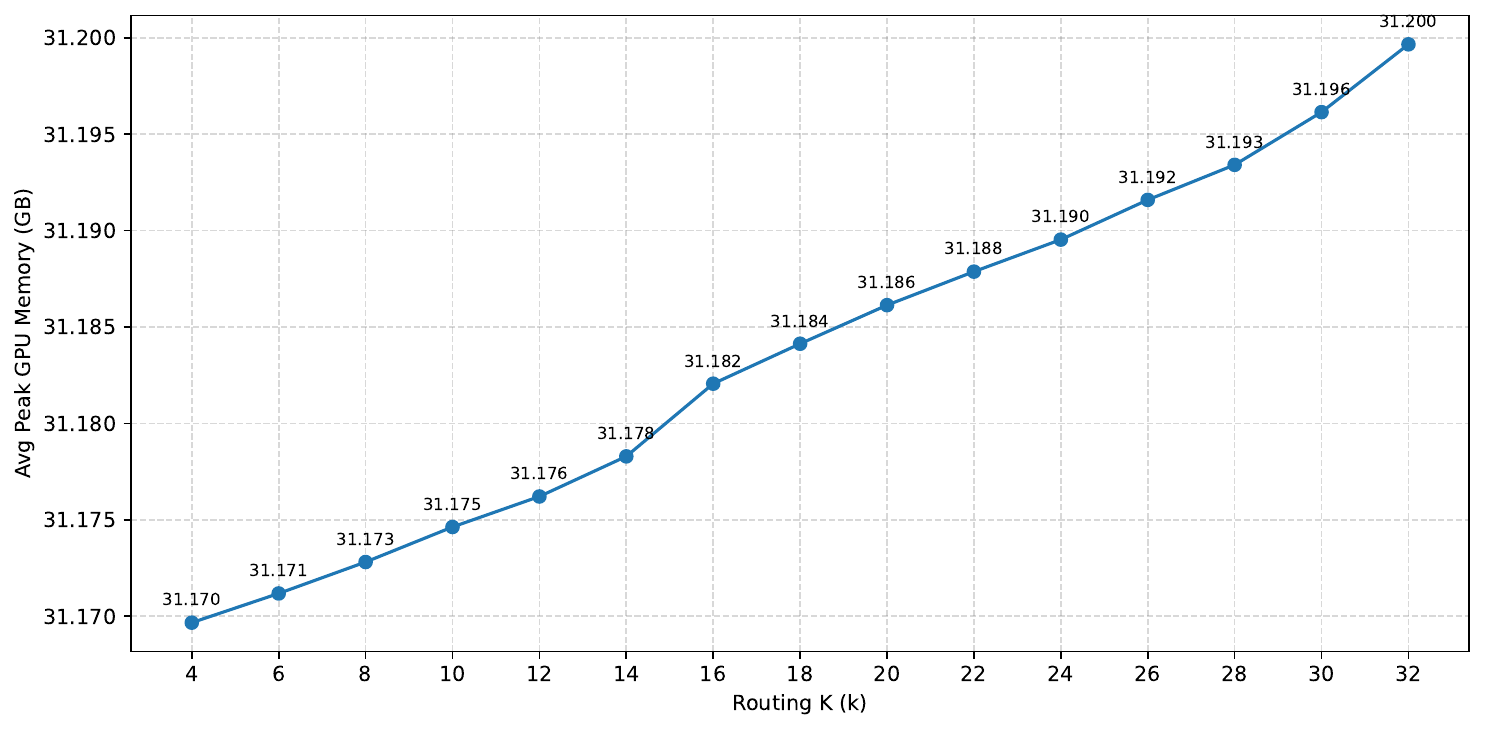}
\caption{Peak GPU memory as a function of the per-parent routing budget \(k\) on NarrativeQA using Qwen2.5-14B-Instruct. Peak memory varies by less than \(0.05\,\mathrm{GB}\) across the sweep.}
\label{fig:routing-memory-sweep}
\end{figure}

\section{LVS dataset and QA construction}
\label{app:lvs}

LVS is a set of VLSI circuit layout verification manuals from a major electronic design automation company.
We derive the LVS-QA set directly from the product manual’s HTML corpus.
Each HTML file is typed as \emph{rich} or \emph{light} using simple structural cues.
We mark a file as \emph{rich} if it contains \texttt{<pre>} or \texttt{<code>} blocks, tables with header rows,
``Warning/Note/Caution'' callouts, option matrices, or numbered constraint lists; otherwise it is labeled \emph{light}.
We reconstruct the document hierarchy from the manual’s table of contents (TOC) and define a file’s \emph{related} context
as its immediate children in this tree.

We instruct an open-source generator, DeepSeek-R1~\cite{deepseek-r1}, to produce one or more question--answer pairs answerable from the file context.
To match file type, we restrict question categories:
\vspace{-2pt}
\begin{itemize}\setlength{\itemsep}{0pt}\setlength{\topsep}{2pt}
\item \textbf{Rich:} definition, usage, syntax, parameter, warning, constraint, list\_children, where\_to\_find\_child, code\_example.
\item \textbf{Light:} definition, scope, usage, list\_children, syntax.
\end{itemize}
\vspace{-2pt}
By construction, each QA item is grounded in the union of a node and its immediately related files in the TOC hierarchy.

\subsection{LVS Results}

\begin{table}[H]
\centering
\small
\setlength{\tabcolsep}{8pt}
\renewcommand{\arraystretch}{1.15}
\caption{ROUGE-L (\%) on LVS using Qwen2.5-14B-Instruct.}
\label{tab:siemens_rougeL}
\begin{tabular}{ccc}
\toprule
Backbone & HMT & H$^{2}$MT \\
\midrule
39.0 & 54.5 & 60.1 \\
\bottomrule
\end{tabular}
\end{table}

\paragraph{Dataset.}
LVS contains 76 technical manuals.
Each manual has, on average, 506K characters and a clear document structure driven by a table of contents (ToC) and nested sections.
We build a global hierarchy with a virtual root, manuals as level-1 nodes, and section nodes organized under their corresponding manual roots.
Since LVS is not publicly available and is highly technical, general-purpose models often exhibit substantial domain mismatch.
We therefore complement the QA-based evaluation (ROUGE-L reported for one manual in the appendix) with an intrinsic language modeling measure. Because LVS is private, we treat it as an appendix-only domain case study. The ROUGE-L result in Table~\ref{tab:siemens_rougeL} is evaluated on one manual; we additionally report perplexity on held-out text and index-construction measurements over 30 manuals.
 
\paragraph{Perplexity evaluation.}
To quantify modeling quality on LVS, we compute test perplexity (PPL) on held-out text.
We compare HMT against \mymodel under the same tokenizer and evaluation script.
Table~\ref{tab:lvs_ppl} shows that \mymodel yields substantially lower PPL, indicating a better fit to the technical distribution.

\begin{table}[H]
\centering
\small

\begin{minipage}{\columnwidth}
\centering
\setlength{\tabcolsep}{6pt}
\renewcommand{\arraystretch}{1.1}
\begin{tabular}{lcc}
\toprule
Method & Test PPL & Relative vs.\ HMT \\
\midrule
HMT & 11.93 & 1.00$\times$ \\
\mymodel & 2.62 & 0.22$\times$ (78.1\% lower; 4.56$\times$ lower) \\
\bottomrule
\end{tabular}
\caption{Perplexity on the LVS test set. Relative values are normalized to HMT.}
\label{tab:lvs_ppl}
\end{minipage}

\vspace{2mm}

\begin{minipage}{\columnwidth}
\centering
\scriptsize
\setlength{\tabcolsep}{6pt}
\renewcommand{\arraystretch}{1.05}
\begin{tabular}{@{}lrrrl@{}}
\toprule
Method & Tree Build Time (s) & Cache (MB) & Dim & Embedder \\
\midrule
\mymodel & 977.97 & 63.06 & 2560 & Qwen3-4B \\
RAPTOR & 5382.67 & 56.96 & 768 & MPNet (multi-qa) \\
\bottomrule
\end{tabular}
\caption{Offline hierarchy and memory construction on 30 LVS manuals. We report embedding dimension (Dim) and the embedding source model (Embedder).}
\label{tab:lvs_eff}
\end{minipage}

\end{table}

\section{Intermediate-node aggregation policies}
\label{app:aggregation}

This appendix provides full definitions of the aggregation function $\mathrm{Agg}$ used in
Section~\ref{sec:mem_emb2}.

\paragraph{Children memory stack.}
Let $v\in V$ be an internal node, let $c_v\triangleq |C(v)|$, and enumerate its children in document order as $C(v)=(u_1,\ldots,u_{c_v})$. We stack their memories row-wise:
\[
M^{\mathrm{ch}}_v
\;\triangleq\;
\begin{bmatrix}
m_{u_1}^{\top}\\
\vdots\\
m_{u_{c_v}}^{\top}
\end{bmatrix}
\in\mathbb{R}^{c_v\times d}.
\]
We then compress $M^{\mathrm{ch}}_v$ into a single vector
\[
\tilde{m}_v \;\triangleq\; \mathrm{Agg}\!\left(M^{\mathrm{ch}}_v\right)\in\mathbb{R}^d.
\]

\paragraph{Aggregation modes.}
Let $M \in \mathbb{R}^{c_v\times d}$ denote the stacked children memory matrix, and $M_i\in\mathbb{R}^d$ its $i$-th row.
Projection matrices map $d\!\to\! d_h$ (head dimension).

\textbf{(1) Mean aggregation}
\[
\mathrm{Agg}_{\mathrm{mean}}(M) \;=\; \frac{1}{c_v}\sum_{i=1}^{c_v} M_i.
\]

\textbf{(2) Self-attention aggregation}
\[
Q = M W_Q,\qquad K = M W_K,
\]
\[
A = \mathrm{softmax}\!\left(\frac{QK^{\top}}{\sqrt{d_h}}\right)\in\mathbb{R}^{c_v\times c_v},
\]
\[
w_i = \sum_{j=1}^{c_v} A_{j i},\qquad \hat{w} = \frac{w}{\sum_{i=1}^{c_v} w_i},
\]
\[
\mathrm{Agg}_{\mathrm{attn}}(M) \;=\; \sum_{i=1}^{c_v} \hat{w}_i\, M_i.
\]

\begingroup
\interdisplaylinepenalty=10000
\paragraph{(3) Parent-conditioned cross-attention aggregation.}
Let \(Z \in \mathbb{R}^{K\times d}\) denote parent-side conditioning tokens used only for aggregation.
\begin{equation*}
\begin{aligned}
Q &= ZW_Q, \qquad K_{\mathrm{key}} = M W_K, \\
S &= \mathrm{softmax}\!\left(\frac{QK_{\mathrm{key}}^{\top}}{\sqrt{d_h}}\right)
     \in \mathbb{R}^{K\times c_v}, \\
w &= \frac{1}{K}\sum_{k=1}^{K} S_{k,:}\in\mathbb{R}^{c_v}, \\
\mathrm{Agg}_{q\text{-}\mathrm{attn}}(M,q) &= \sum_{i=1}^{c_v} w_i\, M_{i,:}.
\end{aligned}
\end{equation*}
\endgroup

\textbf{(4) GAT-style aggregation}
Let $q\in\mathbb{R}^{K\times d}$ be parent query tokens and $\bar{q}\triangleq \frac{1}{K}\sum_{k=1}^{K} q_k\in\mathbb{R}^d$.
\[
k_i = W_{\mathrm{child}} M_i,\qquad p = W_{\mathrm{parent}} \bar{q},
\]
\[
e_i = \mathrm{LeakyReLU}\!\left( a_p^{\top}p + a_c^{\top}k_i \right),
\qquad
\alpha_i = \frac{\exp(e_i/\tau)}{\sum_{j=1}^{c_v}\exp(e_j/\tau)},
\]
\[
\mathrm{Agg}_{\mathrm{GAT}}(M,q) \;=\; \sum_{i=1}^{c_v} \alpha_i \, (W_V M_i).
\]

\textbf{(5) Parent-token self-attention aggregation}
Let $m_{\mathrm{par}}\in\mathbb{R}^{d}$ be a learned parent token and
$\tilde{M}=\mathrm{stack}(\{m_{\mathrm{par}}\}\cup\{M_i\}_{i=1}^{c_v})\in\mathbb{R}^{(1+c_v)\times d}$.
We apply one self-attention layer and take the output at the parent position:
\[
\tilde{Q}=\tilde{M}W_Q,\quad \tilde{K}=\tilde{M}W_K,\quad \tilde{V}=\tilde{M}W_V,
\]
\[
\tilde{A}=\mathrm{softmax}\!\left(\frac{\tilde{Q}\tilde{K}^{\top}}{\sqrt{d_h}}\right)\in\mathbb{R}^{(1+c_v)\times(1+c_v)},
\qquad
\tilde{O}=\tilde{A}\tilde{V},
\]
\[
\mathrm{Agg}_{\mathrm{par\text{-}attn}}(M) \;=\; \tilde{O}_{1,:}\in\mathbb{R}^{d},
\]
where index $1$ denotes the prepended parent position.

\section{Complexity Analysis}
\label{app:complexity}
We decompose inference into (i) offline memory construction on a static corpus and (ii) online query-time routing and answer generation.

\noindent \paragraph{Offline memory construction (amortized).}
Each node $v \in V$ stores a single memory embedding $m_v \in \mathbb{R}^d$ computed bottom-up.
This stage runs once per corpus and is amortized across all queries on the same document.

\noindent\paragraph{Online routing cost.}
At a routed parent $p$, we score each child $u \in C(p)$ by
$s(u \mid q) = (W_q q)^\top (W_k m_u)/\sqrt{d_h}$.
Let $c_p \triangleq |C(p)|$.
Computing scores for all children costs $O(c_p d_h)$ dot-product operations.
With candidate sets $S_0=\{r\}$ and $S_{\ell+1}=\bigcup_{p \in S_\ell} S(p)$ where $S(p)$ returns the top-$k$ children of $p$,
the total routing cost over visited depths is
\[
T_{\mathrm{route}} = O\!\left(\sum_{\ell} \sum_{p \in S_\ell} c_p \, d_h\right).
\]
Under bounded branching $c_p \le c_{\max}$ and depth cap $L_{\max}$, and using $|S_{\ell+1}|\le k|S_\ell|$,
we obtain the upper bound
\[
T_{\mathrm{route}} = O\!\left(c_{\max} d_h \sum_{\ell=0}^{L_{\max}-1} |S_\ell|\right)
= O\!\left(c_{\max} d_h \,\frac{k^{L_{\max}}-1}{k-1}\right).
\]
In practice, traversal stops earlier due to reaching leaves, which caps the number of expanded parents and thus the number of scored children.

\paragraph{Online generation (prefill) cost and the source of speedup.}
We collect selected nodes into a retrieval set $R \subseteq V$ and stack their memories as
$M_{\mathrm{ret}} \triangleq \mathrm{stack}\{m_v : v \in R\}$.
Since each node contributes one embedding, the effective retrieved context length is $|R|$.
Let $n_q$ denote the question length in tokens and let $n_{\mathrm{ret}} \triangleq |R|$.
Standard full-attention Transformer prefill per layer scales as $O(n^2 d)$ for sequence length $n$,
so conditioning on $(M_{\mathrm{ret}}, q)$ costs
\[
T_{\mathrm{prefill}} = O\!\left(L_{\mathrm{LM}} (n_q + n_{\mathrm{ret}})^2 d\right),
\]
where $L_{\mathrm{LM}}$ is the backbone depth.
In contrast, a flat long-context baseline that pre-fills on the entire document of length $N$ incurs
$O(L_{\mathrm{LM}} (n_q + N)^2 d)$.
Therefore, the dominant speedup in TTFT comes from reducing the prefill sequence length from $N$ to $n_{\mathrm{ret}}$,
while paying an additional routing overhead $T_{\mathrm{route}}$ that depends on the explored portion of the hierarchy rather than the raw document length.

\section{Ablation Study: Aggregation Policies}

We compare four of the aggregation policies defined in Appendix~\ref{app:aggregation}: mean aggregation, self-attention aggregation, parent-conditioned cross-attention, and GAT-style aggregation. Table~\ref{tab:openroad-by-model-policy} reports ROUGE-L on OpenROAD, and Table~\ref{tab:openroad-ppl-both} reports corpus perplexity.

\begin{table}[H]
\centering
\scriptsize
\setlength{\tabcolsep}{6pt}
\caption{ROUGE-L (\%) on OpenROAD by model and pooling policy.}
\label{tab:openroad-by-model-policy}
\begin{tabular}{llc}
\toprule
\textbf{Backbone} & \textbf{Aggregation policy} & \textbf{OpenROAD} \\
\midrule
\multirow{4}{*}{\textbf{Qwen2.5-14B}} 
  & Graph-Attention Pooling       & \textbf{32.6} \\
  & Cross-Attention Pooling       & 29.2          \\
  & Mean Pooling                  & 30.1          \\
  & Self-Attention Weight Pooling & 28.5          \\
\midrule
\multirow{4}{*}{\textbf{Llama-3.1-8B}}
  & Graph-Attention Pooling       & 28.8          \\
  & Cross-Attention Pooling       & \textbf{29.0} \\
  & Mean Pooling                  & 27.2          \\
  & Self-Attention Weight Pooling & 27.9          \\
\midrule
\multirow{4}{*}{\textbf{SmolLM-135M}}
  & Graph-Attention Pooling       & \textbf{24.4} \\
  & Cross-Attention Pooling       & 23.0          \\
  & Mean Pooling                  & 22.9          \\
  & Self-Attention Weight Pooling & 23.4          \\
\bottomrule
\end{tabular}
\end{table}

\begin{table}[H]
\centering
\scriptsize
\setlength{\tabcolsep}{6pt}
\caption{Perplexity (PPL) of hierarchical memory construction on the OpenROAD corpus.}
\label{tab:openroad-ppl-both}
\begin{tabular}{lcc}
\toprule
\textbf{Aggregation policy} & \textbf{Qwen2.5-14B} & \textbf{Llama-3.1-8B} \\
\midrule
Graph-Attention Pooling (additive) & \textbf{2.350} & \textbf{2.308} \\
Cross-Attention Pooling            & 2.871          & 3.158 \\
Mean Pooling (baseline)            & 2.815          & 2.647 \\
Self-Attention Weight Pooling      & 3.150          & 3.553 \\
\bottomrule
\end{tabular}
\end{table}

\paragraph{Results and complexity.}
GAT-style aggregation achieves the highest ROUGE-L for Qwen2.5-14B (32.6) and SmolLM-135M (24.4), while parent-conditioned cross-attention is marginally highest for Llama-3.1-8B (29.0 versus 28.8 for GAT-style aggregation). GAT-style aggregation also achieves the lowest perplexity for both Qwen2.5-14B (2.350) and Llama-3.1-8B (2.308). Mean aggregation requires $O(c_vd)$ computation. When the number of parent-conditioning tokens is fixed, parent-conditioned cross-attention and GAT-style aggregation scale linearly with the number of children $c_v$, whereas self-attention aggregation scales quadratically because it compares all pairs of child memories. Thus, no aggregation policy achieves the highest ROUGE-L for every backbone, although GAT-style aggregation obtains the best result in four of the five reported model--metric comparisons.

\section{Training and Inference Configuration}
\label{app:hyper}

\paragraph{Stage 1: corpus training.}
Node memories are constructed in post-order over the semantic hierarchy,
so that the memories of all children are available before constructing
their parent memory. For a leaf node, the backbone processes the node's
local text together with the learnable memory-marker embeddings, and the
final hidden state at the read position forms the node memory. For an
internal node, the selected aggregation module first combines the child
memories into \(\tilde m_v\); the backbone then conditions on
\(\tilde m_v\) and the node's local text to construct the parent memory
\(m_v\). Corpus training optimizes \(\mathcal L_{\mathrm{corpus}}\) and
updates the LoRA adapters, memory-marker embedding, and selected
aggregation module, while the backbone weights remain frozen. The write
and read marker embeddings are tied,
\(e_{\mathrm{write}}=e_{\mathrm{read}}=e_{\mathrm{mem}}\), and
\(e_{\mathrm{mem}}\) is initialized from
\(\mathcal N(0,\sigma_{\mathrm{embed}}^2)\), where
\(\sigma_{\mathrm{embed}}\) is the empirical standard deviation of the
backbone input-embedding matrix. Optimization uses AdamW with learning
rate \(10^{-4}\) and a StepLR schedule with step size \(100\) and
\(\gamma=0.8\).

\paragraph{Stage 2: QA fine-tuning.}
Training resumes from the Stage 1 checkpoint and optimizes
$\mathcal{L}_{\mathrm{QA}}$ for up to 3{,}000 steps with
$\lambda_{\mathrm{ae}} = 0.1$ and StepLR $\gamma = 0.95$. Evidence spans
are aligned to leaf text by string matching and propagated to ancestors to
form per-parent gold sets. During this stage, the cached document-side node memories and aggregation
module are held fixed; the generation, query-encoding, and routing
components are optimized through their corresponding losses.

\paragraph{Parameter-efficient adaptation.}
Backbone weights remain frozen. LoRA uses rank 8, $\alpha = 32$, and
dropout 0.1, applied to the query, key, value, output, gate, up, and down
projections. We exclude the input-embedding matrix from the LoRA target set.

\paragraph{Inference.}
Per-parent routing budgets are $k = 8$ for QASPER and $k = 16$ for
OpenROAD, NarrativeQA, and HotpotQA. Decoding is greedy with a repetition
penalty of 1.2.

\section{Baseline Reproduction and Environment Notes}
\label{app:baseline-configs}

\paragraph{Baseline reproduction.}
For the evaluation in Table~\ref{tab:llama2_7b_nqa_qasper_hpqa}, we
reran the baseline rows using released code or checkpoints under a
single evaluation pipeline. We standardized the dataset split, task
instruction, answer post-processing, context budget, generation
budget, and decoding policy whenever supported by the corresponding
implementation. Method-specific input wrappers and unavoidable
deviations are documented below. Consequently, the baseline values
differ from those reported by the corresponding publications, while
the H$^{2}$MT results are retained from the same LongBench evaluation.
The shared dataset, prompt, context-budget, decoding, and scoring
protocol is specified in Appendix~\ref{app:baseline-reproduction}.

\paragraph{Selective-attention systems.} We additionally attempted to evaluate MagicPIG and IceCache. MagicPIG's LSH kernels require AVX512 intrinsics, which were unavailable on our AMD EPYC hosts. IceCache's Multi-level DCI extension requires OpenBLAS headers that were unavailable on our cluster; substituting MKL produced symbol conflicts with PyTorch. We therefore report RetroInfer and ArkVale, which could be executed in the available environment.

\section{Induced Hierarchy Construction for HotpotQA}
\label{app:gmm-hotpot}

For each HotpotQA document, we split the context into 192-token chunks with 32-token overlap and encode each chunk using the memory constructor. Leaf memories are projected to 10 dimensions using UMAP, after which a Gaussian Mixture Model is fitted independently for that document. The number of mixture components is selected using BIC, and one parent memory is formed for each cluster by aggregating the assigned child memories.

The exploratory extension permits soft assignment: a node is connected to every cluster whose posterior probability exceeds 0.1. The induced structure may therefore be a layered DAG rather than a strict tree. This relaxation is used only for H$^{2}$MT--GMM; all native-structure experiments use rooted trees. During routing, nodes reached through multiple parents are deduplicated before the next level is processed.

The procedure is applied recursively to the newly constructed parent memories. Recursion terminates when the number of clusters is not smaller than the number of input nodes or when depth 8 is reached. The resulting per-document structure is used for top-down routing and memory selection.

\section{Repeated-Query Amortization}
\label{app:amortization}

Let \(T_{\mathrm{off}}^{X}\) and \(T_{\mathrm{on}}^{X}\) denote the one-time construction cost and per-query total generation cost of system \(X\). After \(N\) queries, cumulative cost is \(C^{X}(N)=T_{\mathrm{off}}^{X}+NT_{\mathrm{on}}^{X}\). When H$^{2}$MT has lower construction cost but higher per-query cost than RAPTOR, it remains cheaper while \(N<(T_{\mathrm{off}}^{\mathrm{R}}-T_{\mathrm{off}}^{\mathrm{H}})/(T_{\mathrm{on}}^{\mathrm{H}}-T_{\mathrm{on}}^{\mathrm{R}})\).

On OpenROAD and HotpotQA, both terms favour H$^{2}$MT: hierarchy and memory construction is \(13.5\times\) and \(38.9\times\) faster, respectively, and total generation time is also lower. H$^{2}$MT therefore has lower cumulative wall-clock cost from the first query in these settings.

On QASPER with Qwen3-4B, H$^{2}$MT requires 7.50 seconds of construction per paper compared with 762 seconds for RAPTOR, but its per-query total generation time is approximately 1.09 seconds higher. The crossover occurs at approximately \(N=689\) queries per paper. With Llama-2-7B, the corresponding crossover is approximately \(N=441\). Below these points, H$^{2}$MT has lower cumulative cost; above them, RAPTOR's lower per-query latency dominates. On LVS, construction over 30 manuals takes 978 seconds for H$^{2}$MT and 5,383 seconds for RAPTOR, a \(5.5\times\) reduction.

\section{Query-Encoding Cap}
\label{app:query-cap}

The default query budget is \(n_q=\min(|q|,\lfloor B_{\mathrm{seg}}/2\rfloor)\). We also evaluated a cap of \(B_{\mathrm{seg}}\), corresponding to a ratio-one setting. All evaluated questions were already shorter than the default cap, so the input questions were unchanged and ROUGE-L remained nearly identical. This check confirms that the cap is inactive in the reported experiments rather than demonstrating robustness to actual question truncation.

\section{Implementation Notes}

\paragraph{Training-time relaxation of top-$k$ pruning.}
Although inference uses a hard per-parent $\mathrm{TopK}$ operator, we do not backpropagate through the discrete selection.
Instead, during QA fine-tuning we form a differentiable distribution $\pi_p(u)$ over $u \in C(p)$ by softmaxing the routing scores,
and optimize routing and selection using $\mathcal{L}_{\mathrm{route}}$ and $\mathcal{L}_{\mathrm{sel}}$.
This provides gradients to the routing projections while preserving hard top-$k$ traversal at test time. The cached document-side node memories and the aggregation module are held fixed during QA fine-tuning.

\textbf{Trainable components.}
The backbone weights $\theta$ remain frozen in both stages and are adapted only through LoRA. During corpus training, we optimize the LoRA adapters, the write/read embedding parameters, and the selected aggregation module. During QA fine-tuning, the cached document-side memories and aggregation module remain fixed; the generation loss updates the LoRA adapters, while the routing and selection losses update the routing projections and the trainable query-encoding parameters.

\section{Additional RAPTOR Comparison Analysis}
\label{app:rag-analysis}

\paragraph{Representation trade-off.}
RAPTOR conditions the generator on the explicit text of retrieved nodes, including internal summaries and selected leaf chunks, whereas H$^{2}$MT conditions the reader on fixed-size latent node memories. H$^{2}$MT is therefore intentionally lossy: compressing each node into one bounded representation, aggregating descendant memories, and applying hard top-\(k\) routing can discard fine-grained evidence. This limitation is particularly relevant for multi-hop questions and overlap-based metrics, which benefit from preserving surface-form evidence. RAPTOR retains more explicit textual detail, but its retrieved context can be substantially longer and requires a more expensive offline hierarchy-construction process.

\paragraph{Why RAPTOR has lower TTFT}
RAPTOR embeds each query using a lightweight sentence encoder and performs a single similarity search over precomputed node vectors before reader prefill. H$^{2}$MT instead runs the question through the full LoRA-adapted backbone to construct the routing vector and then performs iterative level-by-level traversal. Its TTFT therefore includes query encoding, candidate-memory loading, projected child scoring, per-parent top-\(k\) selection, and reader prefill.

This routing cost depends on the number of scored candidates and the traversal depth. In the reported experiments, QASPER retains 8.4 node memories on average, whereas OpenROAD retains 26.4, and the larger routing frontier contributes to the higher pre-generation overhead. The routing-budget sweep in Figure~\ref{fig:k_sweep} similarly shows that TTFT increases as \(k\) expands the frontier. These measurements explain why RAPTOR reaches the first token sooner even when it supplies a longer textual context to the generator.

\paragraph{Engineering implications.}
The measured routing overhead is partly implementation-dependent. Keeping node-memory banks resident in device memory, batching candidate loading and scoring across routed parents, or replacing full-backbone query encoding with a lightweight routing encoder could reduce this overhead. These optimizations are not evaluated in the present work.

\subsection{Plain-Text LoRA Ablation}
\label{app:lora-plain-text}

We evaluate whether the LoRA-adapted backbone retains its ordinary
text-based QA behaviour when the H$^{2}$MT memory vectors are replaced
with plain-text context. We evaluate this condition on QASPER
using Qwen3-4B-Instruct, with the H$^{2}$MT LoRA adapter active but
ordinary text supplied instead of latent node memories. This condition
obtains 1.70 ROUGE-L, compared with the substantially stronger
memory-conditioned H$^{2}$MT configuration.

The result indicates that the LoRA adapter specializes to the
latent-memory interface and should not be interpreted as preserving the
checkpoint's ordinary plain-text QA behaviour. The pretrained backbone
weights remain frozen, so the original plain-text model behaviour can be
recovered by disabling the H$^{2}$MT LoRA adapter.

\subsection{Baseline Reproduction Protocol}
\label{app:baseline-reproduction}

\paragraph{HMT reproduction.}
We reproduced HMT from the authors' official implementation and
trained one HMT configuration for each backbone reported in
Tables~\ref{tab:main_rouge} and~\ref{tab:avg_efficiency}. HMT uses
the same backbone checkpoint, LongBench task examples, task
instructions, answer post-processing, generation budgets, and LoRA
fine-tuning protocol as the corresponding H$^{2}$MT configuration.
We use 1,024-token segments and backpropagation through time over
six segments. Optimization uses AdamW with learning rate
\(10^{-4}\). The backbone weights remain frozen, and the model is
fine-tuned using the same LoRA rank, scaling factor, dropout, and
target modules as H$^{2}$MT. We make no algorithmic deviations from
the official HMT implementation.

\paragraph{Models and checkpoints.}
Except for HMT, the baselines in
Table~\ref{tab:llama2_7b_nqa_qasper_hpqa} use publicly released
Llama-2-7B-family checkpoints without additional training or
fine-tuning. HMT is trained from the official implementation as
described above. Table~\ref{tab:baseline-checkpoints} lists the checkpoint and compression mechanism used for each method. AutoCompressors, ICAE, MemoryLLM, and UltraGist use their officially released compression checkpoints or adapters. LongLLMLingua is training-free and applies its released compression procedure using a pretrained Llama-2-7B compressor and a Llama-2-7B-Chat reader. The two MemoryLLM variants use the same checkpoint; MemoryLLM-all-BM25 additionally enables the retrieval mode provided by the official implementation.

\begin{table}[!ht]
\centering
\small
\setlength{\tabcolsep}{3pt}
\renewcommand{\arraystretch}{1.05}
\caption{Baseline sources used in the evaluation. HMT was trained
from the official implementation as described above; the remaining
methods use released checkpoints, adapters, or training-free
procedures.}
\label{tab:baseline-checkpoints}
\begin{tabular}{llll}
\toprule
\textbf{Method} & \textbf{Checkpoint} & \textbf{Backbone} & \textbf{Mechanism} \\
\midrule
Llama-2-7B-Chat
& \texttt{meta-llama/Llama-2-7b-chat-hf}
& Chat
& Uncompressed reader \\
AutoCompressors
& \texttt{princeton-nlp/AutoCompressor-Llama-2-7b-6k}
& Base
& Recursive summary vectors \\
LongLLMLingua
& \texttt{NousResearch/Llama-2-7b-hf} $+$ Llama-2-7B-Chat
& Base $+$ chat
& Training-free token pruning \\
ICAE
& \texttt{getao/icae} adapter on Llama-2-7B-Chat
& Chat
& Memory-slot compression \\
MemoryLLM-16k
& \texttt{YuWangX/memoryllm-7b}
& Base
& Learned memory pool \\
MemoryLLM-all-BM25
& \texttt{YuWangX/memoryllm-7b}
& Base
& Memory pool with BM25 retrieval \\
UltraGist
& \texttt{namespace-Pt/ultragist-llama2-7b-chat}
& Chat
& Fixed-ratio gist compression \\
\bottomrule
\end{tabular}
\end{table}

\paragraph{Shared evaluation protocol.}
We evaluate all methods under a common inference pipeline.
Each example is limited to at most 4,096 source tokens before
method-specific processing. Inputs exceeding this limit are
truncated using the LongBench middle-truncation strategy,
which retains tokens from the beginning and end of the
document. All methods use greedy decoding with batch size
one, \texttt{do\_sample=False}, \texttt{num\_beams=1}, and a
repetition penalty of 1.2. We allow at most 128 generated
tokens for QASPER and 64 tokens for HotpotQA and NarrativeQA.
The same task instruction and answer post-processing
procedure are used across methods. Chat checkpoints receive
the corresponding chat template, whereas base checkpoints
receive a plain-text prompt, following the formatting
expected by their official implementations.

\paragraph{Datasets and metrics.}
For QASPER, HotpotQA, and NarrativeQA, we use the 200-example
evaluation sets distributed with LongBench. The H$^{2}$MT results
and all rerun baseline rows in
Table~\ref{tab:llama2_7b_nqa_qasper_hpqa} use the same
task-specific LongBench examples. The table reports token-level
F1, taking the maximum score over the available reference answers.
We additionally record the input-token count, time to the first
generated token, and peak accelerator memory when these
measurements are supported by the released implementation.

\paragraph{Hardware.}
All Table~\ref{tab:llama2_7b_nqa_qasper_hpqa} baseline evaluations
are performed on a single AMD Instinct MI210 GPU using bfloat16
precision and scaled dot-product attention.
\paragraph{Method-specific settings.}
AutoCompressors uses 50 summary vectors with recursive compression. MemoryLLM-all-BM25 retrieves over 512-token source chunks using the retrieval mode in the official implementation. ICAE is designed to compress one 512-token segment into 128 memory slots. We therefore report its native 512-token operating point as the primary ICAE result. Time-to-first-token is omitted where it could not be measured consistently through the method's custom decoding implementation.

\subsection{Dataset-Specific Prompt Templates}
\label{app:prompt-templates}

We use dataset-specific instructions rather than a single prompt shared
across all benchmarks. Within each dataset and evaluation setting, the
same instruction and answer post-processing are used for all compared
methods. Chat-based prompts are serialized using the corresponding
checkpoint's native chat template. Table~\ref{tab:prompt-templates}
reports the exact instruction used in each evaluation path.

\begin{table}[!ht]
\caption{Dataset-specific instructions used in evaluation. The question
placeholder is denoted by \texttt{\{q\}}.}
\label{tab:prompt-templates}
\centering
\scriptsize
\setlength{\tabcolsep}{4pt}
\begin{tabular}{lp{0.58\textwidth}l}
\toprule
Dataset / setting & Instruction & Serialization \\
\midrule
QASPER
&
\texttt{Prompt: Answer the question based on the provided scientific
context. Question: \{q\} Answer:}
&
Plain text
\\

HotpotQA
&
\texttt{Answer the following multi-hop question.}
&
Chat system message
\\

HotpotQA, GMM
&
\texttt{Answer the following multi-hop question.}
&
Chat system message
\\

NarrativeQA
&
\texttt{Answer the following question based on the provided document.}
&
Chat system message
\\

LVS / Calibre
&
\texttt{Answer the following question about Calibre documentation.}
&
Chat system message
\\

OpenROAD
&
\texttt{You are an expert with EDA tool usage. Answer the question based
on the following reference information.}
&
Included in processed input
\\
\bottomrule
\end{tabular}
\end{table}

\end{document}